\documentclass{article}

\PassOptionsToPackage{table}{xcolor}
\usepackage{iclr2027_conference,times}
\iclrfinalcopy

\usepackage[utf8]{inputenc}
\usepackage[T1]{fontenc}
\usepackage{microtype}
\usepackage{xcolor}
\usepackage{graphicx}
\usepackage{booktabs}
\usepackage{multirow}
\usepackage{enumitem}
\usepackage{array}
\usepackage{ragged2e}
\usepackage{amsmath}
\usepackage{amsfonts}
\usepackage{adjustbox}
\usepackage{wrapfig}
\usepackage{pifont}
\usepackage{xspace}
\usepackage{tcolorbox}
\tcbuselibrary{breakable,skins}
\usepackage{fontawesome5}
\usepackage{hyperref}
\usepackage[nameinlink,noabbrev]{cleveref}

\definecolor{tealblue}{RGB}{16, 102, 102}
\definecolor{brightteal}{RGB}{45, 170, 165}
\definecolor{headerbg}{RGB}{240, 242, 245}
\definecolor{goodgreen}{RGB}{40, 140, 60}
\definecolor{badred}{RGB}{200, 50, 50}
\hypersetup{
  colorlinks=true,
  linkcolor=tealblue,
  citecolor=brightteal,
  urlcolor=brightteal
}
\newcommand{\cmark}{\textcolor{goodgreen}{\ding{51}}}
\newcommand{\xmark}{\textcolor{badred}{\ding{55}}}

\newcommand{\ourmethod}{JIT-Agent\xspace}
\definecolor{toponecell}{RGB}{211,234,214}
\definecolor{toptwocell}{RGB}{232,241,204}
\definecolor{topthreecell}{RGB}{248,241,203}
\newcommand{\topone}[1]{\cellcolor{toponecell}\textbf{#1}}
\newcommand{\toptwo}[1]{\cellcolor{toptwocell}#1}
\newcommand{\topthree}[1]{\cellcolor{topthreecell}#1}

\newcommand{\ghlink}{\url{https://github.com/bingreeky/JIT}}
\newcommand{\githubicon}{\raisebox{-1.2pt}{\includegraphics[height=1em]{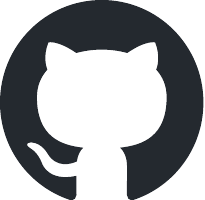}}}

\newtcolorbox{insightbox}[1]{
  enhanced jigsaw,
  breakable,
  colback=tealblue!4!white,
  colframe=tealblue!82!black,
  boxrule=0.8pt,
  arc=4pt,
  left=7pt,
  right=7pt,
  top=6pt,
  bottom=4pt,
  before skip=9pt,
  after skip=9pt,
  fontupper=\small,
  title={#1},
  colbacktitle=tealblue,
  coltitle=white,
  fonttitle=\scriptsize\bfseries,
  attach boxed title to top left={xshift=12pt,yshift=-2.6mm},
  boxed title style={
    colback=tealblue,
    colframe=tealblue,
    boxrule=0pt,
    arc=2pt,
    left=6pt,
    right=6pt,
    top=2pt,
    bottom=2pt
  }
}
\title{JIT-Agent: Scaling Harness Intelligence via Just-in-Time Harness Evolution}
\author{JIT-Agent Team
}
\begin{document}

\maketitle
\vspace{-1.2em}
\begin{abstract}
Agent capability is not determined by the model alone. The agent harness, encompassing memory management, planning strategy, action protocol, and tool/skill orchestration, can dominate the contribution of the underlying foundation model. Yet harness design remains manual, task-specific, and fundamentally unscalable. We present \ourmethod, a harness intelligence model trained to synthesize task-adaptive agent harnesses on the fly for arbitrary off-the-shelf agentic LLMs. We formalize the agent harness as a composable, machine-generatable artifact governed by a fixed four-module protocol, and train \ourmethod to \textit{customize harnesses} for a given task at hand, \textit{repair harnesses} for stable and reliable execution, and \textit{self-evolve} by distilling performance signals from an expanding archive of prior harness configurations. Equipped with \ourmethod as a harness helper, DeepSeek-V4-Flash surpasses GPT-5.6 on DeepSearchQA ($+9.1$) and OdysseyBench ($+4.3$), while the already strong GLM-5.2 gains up to $+20.2$ points. Across controlled evaluations, \ourmethod-generated harnesses are performance-competitive with mature agent runtimes such as OpenCode and Claude Code and consistently improve multi-scale model families of DeepSeek V4, Mimo-V2.5, and Qwen3.6. To our knowledge, \ourmethod is the first model purpose-built for just-in-time harness generation, establishing harness intelligence as a trainable, transferable, and compounding dimension of agent capability orthogonal to model scaling. \\[0.5em]\githubicon\;Code: \ghlink
\end{abstract}


\vspace{-0.5em}
\section{Introduction}
\vspace{-0.5em}

\label{sec:introduction}
{Agent harness matters.}
The capability of an LLM agent is jointly determined by two tightly coupled factors: the \textit{foundation model} that produces reasoning and actions, and the \textit{agent harness} that situates this model in a closed-loop execution environment~\citep{meng2026agent,zhou2026externalizationllmagentsunified,ning2026code}.
The harness decides what history is retained, how local intent is formed, which tools or skills are exposed, how actions are executed, and when verification or recovery is triggered~\citep{meng2026agent,zhou2026externalizationllmagentsunified}.
A strong model may fail when placed behind the wrong memory, planner, or action protocol; a strong harness, in turn, can only unlock capability when the model can understand and follow it~\citep{lee2026metaharnessendtoendoptimizationmodel,lin2026agenticharnessengineeringobservabilitydriven}.
Thus, agent intelligence is not a property of model weights alone, but of the model--harness pair~\citep{lee2026metaharnessendtoendoptimizationmodel,zhou2026externalizationllmagentsunified}.
Harness design is therefore not an implementation detail, but a first-order determinant of agent performance~\citep{meng2026agent,lin2026agenticharnessengineeringobservabilitydriven}.

\vspace{-0.5em}
\paragraph{Ahead-of-Time (AOT) Harness} This observation has motivated a growing line of work on test-time harness optimization~\citep{lee2026metaharnessendtoendoptimizationmodel,lin2026agenticharnessengineeringobservabilitydriven}.
Recent systems optimize harness code, prompts, tools, memories, skills, or control policies from trajectories and feedback, yielding strong improvements in coding, terminal-use, web interaction, and general agentic domains~\citep{lee2026metaharnessendtoendoptimizationmodel,lin2026agenticharnessengineeringobservabilitydriven}.
Despite their differences, many of these methods share an \textit{Ahead-of-Time} (AOT) assumption: the harness is treated as a durable artifact to be optimized over an experience stream, with the hope that the resulting artifact will generalize across future tasks, domains, or model versions~\citep{lee2026metaharnessendtoendoptimizationmodel,lin2026agenticharnessengineeringobservabilitydriven}.
This is a powerful paradigm when the deployment distribution is stable and homogeneous~\citep{lin2026agenticharnessengineeringobservabilitydriven,zhang2025memevolve}.
However, it still asks the optimization loop to precompile a broadly useful harness before seeing the exact structure of each future problem~\citep{lee2026metaharnessendtoendoptimizationmodel}.

\vspace{-0.5em}
\paragraph{Just-in-Time (JIT) Harness.} We instead turn to a \textbf{\textit{Just-in-Time} (JIT) view of harness construction}.
Different tasks plainly require different harness priors.
Wide-search tasks may benefit from parallel evidence exploration~\citep{qin2025flashsearcherfasteffectiveweb}; terminal tasks often favor a lean serial ReAct loop~\citep{yao2023reactsynergizingreasoningacting,merrill2026terminalbenchbenchmarkingagentshard}; deep-research tasks require working memory over retrieved evidence~\citep{hu2025memoryageaiagentssurvey}; and NL2Repo-style coding tasks are naturally mediated by a filesystem that stores patches, tests, traces, and repository state~\citep{yang2024sweagentagentcomputerinterfacesenable,ning2026code}.
In short, the appropriate harness is not only domain-dependent, but instance-dependent~\citep{meng2026agent,zhou2026externalizationllmagentsunified}.
Optimizing a single AOT harness across such heterogeneous demands is cumbersome: one must search a large design space, accumulate enough trajectories, and then hope that the resulting scaffold matches the next task~\citep{lee2026metaharnessendtoendoptimizationmodel,lin2026agenticharnessengineeringobservabilitydriven}.
We ask whether another possibility is viable: \textbf{Model-as-a-Harness}, where a trained meta-agent generates a task-specific harness on the spot, and an arbitrary agentic LLM executes under that harness.

\begin{figure}[!t]
    \centering
    \includegraphics[width=\textwidth]{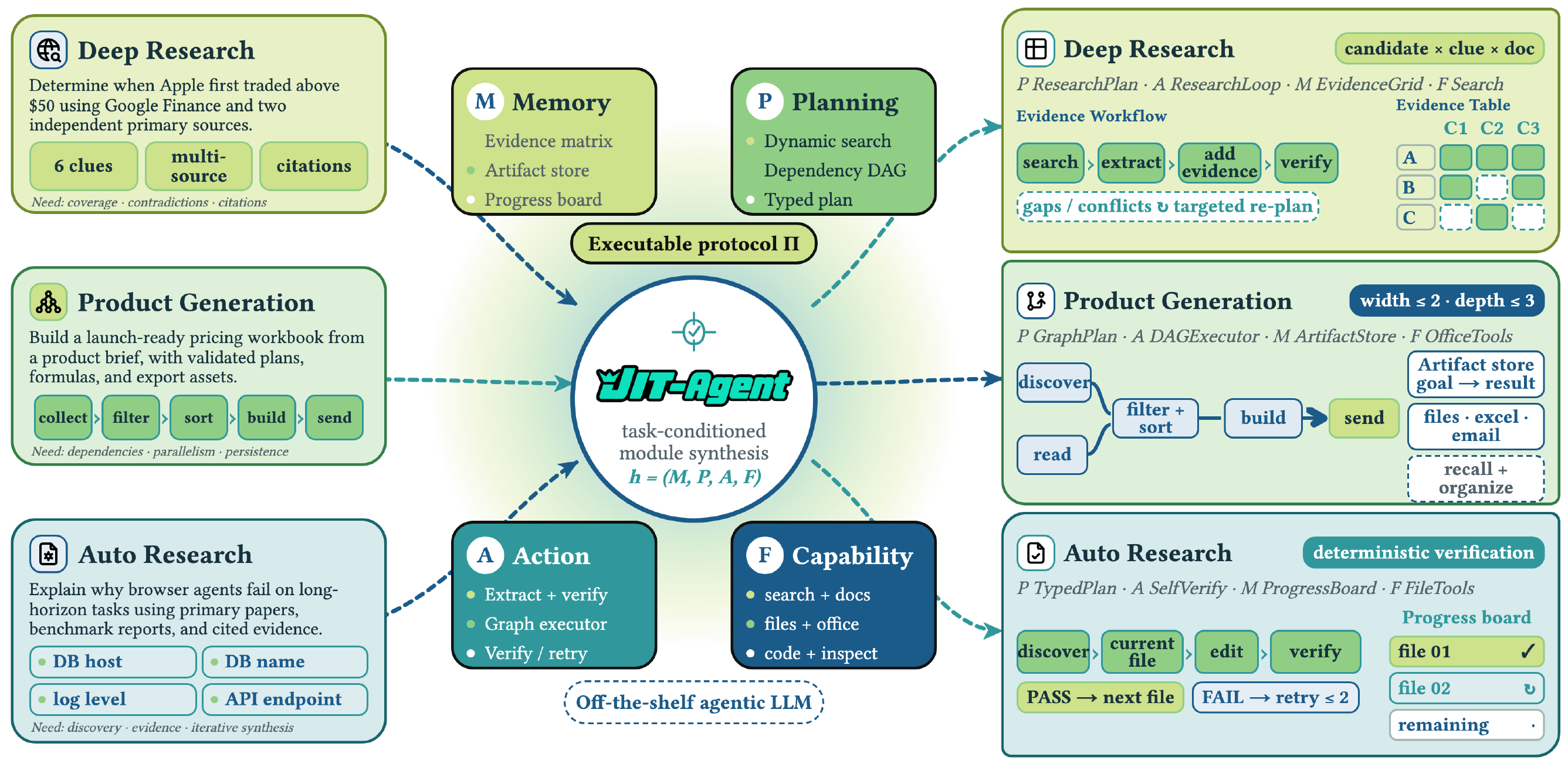}
    \caption{\textbf{Overview of JIT-Agent.}
Given a task, JIT-Agent composes a problem-specific agent harness by instantiating (rather than simply combining) four modules: memory, planning, action, and capability. Different task structures therefore induce distinct executable protocols and state organizations, as illustrated by the specialized harnesses for deep research, product generation, and autonomous research.}
    \label{fig:overview}
\end{figure}

\vspace{-0.5em}
\paragraph{JIT-Agent.} We propose \ourmethod, a compact meta-agent for just-in-time harness generation and evolution.
At inference time, \ourmethod{} receives a task specification, a protocol, an executable tool/skill registry, and a small retrieved context of prior harnesses, and emits an executable harness tailored to the task at hand.
The resulting harness wraps an off-the-shelf agentic LLM for execution. As feedback and traces accumulate, \ourmethod{} can revise the harness and update a harness archive, enabling test-time harness evolution while the generator itself remains fixed.
Training such a system raises three challenges:
\ding{182} \textbf{adaptivity}, namely how to match the generated harness to the task;
\ding{183} \textbf{reliability}, namely how to ensure executable behavior and recover when synthesis fails~\citep{pan2026natural,ning2026code}; and
\ding{184} \textbf{evolvability}, namely how to turn execution feedback into stronger future harnesses~\citep{lee2026metaharnessendtoendoptimizationmodel,lin2026agenticharnessengineeringobservabilitydriven}.
Together, these three requirements define what we call \textbf{harness intelligence}:

\begin{insightbox}{Harness Intelligence}
Harness intelligence is the capacity to construct and refine the operational scaffold through which a model acts. It has three defining properties: \textbf{adaptivity}, matching the harness to the task and backbone; \textbf{reliability}, producing executable behavior and recovering when synthesis fails; and \textbf{evolvability}, turning execution feedback into stronger future harnesses. Intelligence about the harness begins where competence within a fixed harness ends.
\end{insightbox}

\ourmethod realizes harness intelligence through a three-stage training recipe over a protocol-induced harness space~\citep{meng2026agent,zhou2026externalizationllmagentsunified}.
First, we factor an agent harness into a four-tuple of \textit{memory}, \textit{planning}, \textit{action}, and \textit{capability-orchestration} modules, and build \textsc{HarnessFactory}, a unified codebase of representative scaffolds under this common interface~\citep{hu2025memoryageaiagentssurvey,erdogan2025plan,yao2023reactsynergizingreasoningacting,xu2026agentskillslargelanguage}.
Following the view that \textit{code carries the harness}~\citep{ning2026code,pan2026natural}, this representation makes harness generation feasible: \ourmethod{} generates structured executable modules rather than unconstrained agent programs.
\textbf{\ding{182} Stage I} then teaches task-conditioned customization from teacher-generated protocol-compliant examples, giving the generator adaptivity over the shared design space.
Second, \textbf{\ding{183} Stage II} converts failed generations into bounded repair trajectories, turning compiler errors, interface mismatches, and runtime failures into supervision for reliable recovery.
Third, \textbf{\ding{184} Stage III} introduces \textit{Evolutionary Group-Decoupled Policy Optimization} (Evo-GDPO), which optimizes \ourmethod{} to propose harnesses that overtake the current archive frontier while separately normalizing reward, latency, and cost.
This supplies evolvability by turning test-time harness improvement into a trained capability rather than an external search heuristic.
Through this pipeline, we train \textbf{JIT-Agent-27B}, a \textbf{harness intelligence model} that can customize, repair, and evolve task-specific harnesses, shifting harness engineering from AOT artifacts to native JIT synthesis.

\vspace{-0.5em}
\paragraph{Experiment Findings.}
Across nine benchmarks, \ourmethod consistently strengthens its underlying backbones: DeepSeek-V4-Flash surpasses GPT-5.6 by $9.1$ points on DeepSearchQA and $4.3$ on OdysseyBench, while GLM-5.2 gains up to $20.2$ points. The generated harnesses also remain competitive with mature runtimes and transfer across DeepSeek V4, Mimo-V2.5, and Qwen3.6.

\vspace{-0.5em}
\paragraph{Contributions.} (I) We formulate \textit{Model-as-a-Harness} and define \textit{harness intelligence} as the learned ability to construct, repair, and evolve an agent's operational scaffold. (II) We instantiate this paradigm with \ourmethod, which combines a four-module executable protocol, \textsc{HarnessFactory}, and a three-stage customization--repair--evolution pipeline, and (III) validate that JIT-generated harnesses improve diverse backbones while remaining competitive with strong production harnesses.

\vspace{-0.5em}
\section{Related Work}
\vspace{-0.5em}
\paragraph{Harness Engineering and Modularization.} Harnesses have diversified across coding and general-purpose runtimes such as Claude Code, Codex, Hermes, and OpenClaw~\citep{githubGitHubAnthropicsclaudecode,openaiCodexCoding,nous2026hermes,openclaw2026}, embodied systems~\citep{zhang2026harnessvla,lee2026physicalaiharness}, search agents~\citep{apodex2026,qin2025flashsearcherfasteffectiveweb}, and memory-centric systems~\citep{chhikara2025mem0,hu2026evermemos,zhang2025memevolve}. Despite domain-specific implementations, they share an operational layer that governs memory, planning and control, action verification, and tool/skill orchestration, making capability a property of the model--harness pair~\citep{meng2026agent,zhou2026externalizationllmagentsunified,pan2026natural}. Modular representations expose these choices through explicit interfaces, allowing components to be substituted and optimized independently~\citep{chen2026harnessx,ning2026code}. We therefore use the compact executable factorization $\mathbf{h}=(\mathbf{M},\mathbf{P},\mathbf{A},\mathbf{F})$ for memory, planning, action, and capability orchestration.
\vspace{-0.5em}
\paragraph{Harness Optimization.} With explicit interfaces, harness construction becomes optimization over executable artifacts rather than a fixed engineering choice~\citep{khattab2023dspy,yuksekgonul2024textgradautomaticdifferentiationtext,hu2024adas,shang2024agentsquare}. Recent methods extend optimization from prompts and workflow graphs to complete harness code, commonly using a diagnose--propose--validate loop in which traces guide edits and validation determines retention~\citep{zhang_aflow_2024,lee2026metaharnessendtoendoptimizationmodel,lin2026agenticharnessengineeringobservabilitydriven,chen2026harnessx}. Recursive, hierarchical, and online variants improve this loop, but still primarily optimize a persistent artifact against accumulated experience and are therefore ahead-of-time relative to a new task instance~\citep{lee2026recursiveharness,zhou2026hierarchicalselfimprovement,nie2026tthe,shao2026harnessr1}. \ourmethod instead amortizes search into a trained generator that constructs a task-conditioned harness just in time; \Cref{tab:harness_optimization_comparison} summarizes this distinction.
\vspace{-0.5em}
\section{Unified Harness Codebase}
\vspace{-0.5em}
\label{sec:harness_codebase}

\vspace{-0.5em}
\subsection{Modularized Harness Design Space}
\vspace{-0.5em}
An agent harness is the operational layer that turns a foundation model into a closed-loop agent~\citep{202604.0428,zhou2026externalizationllmagentsunified}: it determines what from prior interaction is retained~\citep{hu2026memoryageaiagents}, how intermediate directives are formed~\citep{cao2025largelanguagemodelsplanning}, which external capabilities are exposed at each stage~\citep{xu2026agentskillslargelanguage,yang2026efficientagentsmemorytool}, and how control advances. Harness generation is therefore defined over programs rather than unconstrained text. We distinguish the raw generation space $\mathcal{G}$, syntactically valid harnesses $\mathcal{H}^{\mathrm{syn}}$, protocol-compliant harnesses $\mathcal{H}_{\boldsymbol{\Pi}}$, and their executable subset $\mathcal{H}^{\mathrm{exec}}_{\boldsymbol{\Pi}}$, with $\mathcal{G}\supseteq\mathcal{H}^{\mathrm{syn}}\supseteq\mathcal{H}_{\boldsymbol{\Pi}}\supseteq\mathcal{H}^{\mathrm{exec}}_{\boldsymbol{\Pi}}$. The fixed protocol $\boldsymbol{\Pi}$ specifies module schemas and interfaces, their lifecycle, validation rules, and the shared execution semantics. It removes incidental variation in language and runtime while retaining the principal operational choices that distinguish existing harnesses~\citep{autogen,githubGitHubAg2aiag2,githubGitHubLangchainailanggraph,luo2025largelanguagemodelagent}.
\ourmethod makes these choices explicit through four interoperable modules: \textit{how history is compressed}, \textit{how local intent is formed}, \textit{how tools and skills are orchestrated}, and \textit{how control advances}. Let $\boldsymbol{\tau}$ be a task, $\pi_{\psi}$ a frozen backbone executor, and $\mathcal{C}_{\tau}$ the capability registry available (\textit{e.g.}, tools, APIs, skills) to that task. Running a harness $\mathbf{h}$ with $\pi_{\psi}$ induces a closed-loop trajectory
\begin{equation}
\label{eq:trajectory}
    \boldsymbol{\xi}
    \sim \operatorname{Rollout}(\boldsymbol{\tau},\pi_{\psi},\mathbf{h},\mathcal{C}_{\tau};\boldsymbol{\Pi})
    = \bigl(\mathbf{s}_1,e_1,\mathbf{o}_1,\ldots,\mathbf{s}_T,e_T,\mathbf{o}_T\bigr),
\end{equation}
where $\mathbf{s}_t\in\mathcal{S}$ is the maintained controller state, $e_t\in\mathcal{A}=\mathcal{U}\sqcup\mathcal{Y}$ is either an executable call from $\mathcal{U}$ or a terminal output from $\mathcal{Y}$, $\mathbf{o}_t\in\mathcal{O}$ is the resulting observation, and $T$ is the protocol- or budget-bounded stopping time. Our key assumption is that every $\mathbf{h}\in\mathcal{H}_{\boldsymbol{\Pi}}$ admits:
\begin{equation}
\label{eq:harness_decomp}
    \mathbf{h} \,=\, \bigl(\mathbf{M},\, \mathbf{P},\, \mathbf{A},\, \mathbf{F}\bigr)
    \in
    \mathfrak{M} \times \mathfrak{P} \times \mathfrak{A} \times \mathfrak{F},
\end{equation}
where $\mathbf{M}$, $\mathbf{P}$, $\mathbf{A}$, and $\mathbf{F}$ denote memory, planning, action, and capability-orchestration modules, and $\mathfrak{M}$, $\mathfrak{P}$, $\mathfrak{A}$, and $\mathfrak{F}$ are their protocol-compatible implementation spaces. The tuple follows this conceptual decomposition; at runtime, its dependency order is $\mathbf{M}\!\rightarrow\!\mathbf{P}\!\rightarrow\!\mathbf{F}\!\rightarrow\!\mathbf{A}$. All modules operate against the same frozen $\pi_{\psi}$, whose common runtime dependence is suppressed below.
The protocol maintains both the immutable event history $\boldsymbol{\xi}_{<t}$ and the mutable controller state $\mathbf{s}_t$, which interact as
\begin{equation}
\label{eq:module_dynamics}
\begin{aligned}
    \mathbf{v}_t &= \mathbf{M}(\boldsymbol{\xi}_{<t},\mathbf{s}_t) \in \mathcal{V},
    && \text{history $\to$ view,} \\
    \mathbf{d}_t &= \mathbf{P}(\boldsymbol{\tau},\mathbf{s}_t,\mathbf{v}_t) \in \mathcal{D}_{\mathrm{dir}},
    && \text{view $\to$ local directive,} \\
    \mathcal{C}_t &= \mathbf{F}(\mathcal{C}_{\tau},\mathbf{s}_t,\mathbf{v}_t,\mathbf{d}_t) \subseteq \mathcal{C}_{\tau},
    && \text{directive-conditioned capability orchestration,} \\
    (\mathbf{s}_{t+1},e_t) &= \mathbf{A}(\mathbf{s}_t,\boldsymbol{\tau},\mathbf{v}_t,\mathbf{d}_t,\mathcal{C}_t) \in \mathcal{S}\times\mathcal{A},
    && \text{control update and action emission,}
\end{aligned}
\end{equation}
where $\mathcal{V}$ and $\mathcal{D}_{\mathrm{dir}}$ are the view and directive spaces. The registry $\mathcal{C}_{\tau}$ contains callable tools (\textit{e.g.}, bash tools, APIs, and MCPs~\citep{sarkar2025surveyllmagentcommunication}) as well as higher-level agent skills~\citep{xu2026agentskillslargelanguage,claudeAgentSkills}; a harness without an explicit planner remains type-consistent through the null directive $\mathbf{d}_{\emptyset}\in\mathcal{D}_{\mathrm{dir}}$ returned by $\mathbf{P}_{\emptyset}$. Thus, $\mathbf{M}$ constructs a view of realized history, $\mathbf{P}$ forms a local directive, $\mathbf{F}$ activates relevant capabilities, and $\mathbf{A}$ updates the controller state and emits the next action. The shared kernel $\operatorname{Exec}:\mathcal{U}\times2^{\mathcal{C}_{\tau}}\rightarrow\mathcal{O}$ sets $\mathbf{o}_t=\operatorname{Exec}(e_t;\mathcal{C}_t)$ for $e_t\in\mathcal{U}$ and $\mathbf{o}_t=\bot$ for $e_t\in\mathcal{Y}$, where $\bot\in\mathcal{O}$ is the terminal null observation; it then updates $\boldsymbol{\xi}_{\le t}=\boldsymbol{\xi}_{<t}\oplus(\mathbf{s}_t,e_t,\mathbf{o}_t)$, with $\oplus$ denoting event append. Execution starts from $\boldsymbol{\xi}_{<1}=\emptyset$ and a protocol-defined $\mathbf{s}_1\in\mathcal{S}$, and terminates when $e_t\in\mathcal{Y}$, yielding $y=e_T\in\mathcal{Y}$. This factorization maps otherwise heterogeneous programs into comparable coordinates in $\mathfrak{M}\times\mathfrak{P}\times\mathfrak{A}\times\mathfrak{F}$: canonical ReAct uses full-history memory, no explicit planner, a ReAct loop, and the full capability registry; engineered variants such as Codex and OpenCode add compact memory and todo-based planning~\citep{openaiCodexCoding,githubGitHubOpencodeaiopencode}; and recursive systems such as ROMA, AOrchestra, and RLM instantiate subproblem memory, decomposition, recursive action, and routed capabilities~\citep{alzubi2026romarecursiveopenmetaagent,ruan2026aorchestraautomatingsubagentcreation,zhang2026recursivelanguagemodels}. 
Given this modularized design space, we next introduce \textsc{HarnessFactory}, whose role is to test the expressiveness of $\mathcal{H}_{\boldsymbol{\Pi}}$ and furnish diverse source material for \ourmethod's meta-harness design.

\begin{table}[!t]
\setcitestyle{numbers,square}
\centering
\caption{Seed bank $\mathcal{B}_0$: 13 hand-written harnesses instantiating the four-module protocol $\boldsymbol{\Pi}$. Each row is a complete harness; columns follow the conceptual order of memory, planning, action, and capability orchestration.}
\label{tab:seed_harnesses}
\setlength{\tabcolsep}{5pt}
\renewcommand{\arraystretch}{0.9}
\resizebox{\textwidth}{!}{\begin{tabular}{lllll}
\toprule
\rowcolor{headerbg}
    & \textcolor{tealblue}{\ding{168}}~\textbf{Memory}
    & \textcolor{tealblue}{\ding{169}}~\textbf{Planning}
    & \textcolor{tealblue}{\ding{170}}~\textbf{Action}
    & \textcolor{tealblue}{\ding{171}}~\textbf{Capability Orchestration} \\
\rowcolor{headerbg} \multirow{-2}{*}{\textbf{Harness}}
    & \scriptsize $\mathbf{M}\in\mathfrak{M}$
    & \scriptsize $\mathbf{P}\in\mathfrak{P}$
    & \scriptsize $\mathbf{A}\in\mathfrak{A}$
    & \scriptsize $\mathbf{F}\in\mathfrak{F}$ \\
\midrule
\textsc{ReAct}~\citep{yao2023react}
    & FullHistory
    & No explicit planner
    & ReAct
    & Full registry
\\
\textsc{Plan-and-Execute}~\citep{erdogan2025planandactimprovingplanningagents}
    & FullHistory
    & Linear roadmap
    & ReAct
    & Full registry
\\
\textsc{ReSum}~\citep{wu2026resumunlockinglonghorizonsearch}
    & ReSum memory
    & No explicit planner
    & ReAct
    & Full registry
\\
\textsc{Flash-Searcher}~\citep{qin2025flashsearcherfasteffectiveweb}
    & FullHistory
    & DAG planning
    & ReAct
    & Full registry
\\
\textsc{GAM}~\citep{yan2025generalagenticmemorydeep}
    & GAM retrieval
    & DAG
    & ReAct
    & Full registry
\\
\textsc{MemoBrain}~\citep{qian2026memobrain}
    & Reasoning graph
    & No explicit planner
    & Marker-guided execution
    & Full registry
\\
\textsc{AggAgent}~\citep{lee2026agenticaggregation}
    & Isolated rollout histories
    & No explicit planner
    & Multi-rollout aggregation
    & Full registry
\\
\textsc{OAgent}~\citep{zhu2025oagentsempiricalstudybuilding}
    & Coordinator history
    & No explicit planner
    & Ensemble voting
    & Full registry
\\
\textsc{AgentFold}~\citep{ye2025agentfoldlonghorizonwebagents}
    & AgentFold memory
    & DAG
    & ReActFold
    & Full registry
\\
\textsc{HiAgent}~\citep{hu2024hiagenthierarchicalworkingmemory}
    & Hierarchical memory
    & No explicit planner
    & ReAct
    & Full registry
\\
\textsc{DeepAgent}~\citep{li2026deepagentgeneralreasoningagent}
    & Three-tier memory
    & No explicit planner
    & Marker-guided execution
    & Tool search
\\
\textsc{ROMA}~\citep{alzubi2026romarecursiveopenmetaagent}
    & Context isolation
    & Atomizer + DAG
    & Recursive execution
    & Full registry
\\
\textsc{AOrchestra}~\citep{ruan2026aorchestraautomatingsubagentcreation}
    & Context isolation
    & Atomizer + DAG
    & Recursive execution
    & Agent Delegation
\\
\bottomrule
\end{tabular}}
\end{table}

\vspace{-0.5em}
\subsection{HarnessFactory}
\vspace{-0.5em}
Under the shared protocol $\boldsymbol{\Pi}$ and a common kernel, \textsc{HarnessFactory} re-implements 13 representative scaffolds spanning memory, planning, aggregation, hierarchical control, and recursive execution; \Cref{tab:seed_harnesses} gives the full inventory. These implementations form the protocol-compatible seed bank $\mathcal{B}_0$ of size $K_0=13$. The bank serves a dual role: its seeds anchor Stage-I synthesis, while later states supply the prior population against which new designs are evaluated. After $n$ evolution rounds, $\mathcal{B}_0$ grows into $\mathcal{B}_n\supseteq\mathcal{B}_0$, whose entries associate each retained harness with its task and observed reward, latency, and cost; this evolution is formalized in \Cref{sec:harness_evolution}.

\begin{figure}[!t]
    \centering
    \includegraphics[width=\linewidth]{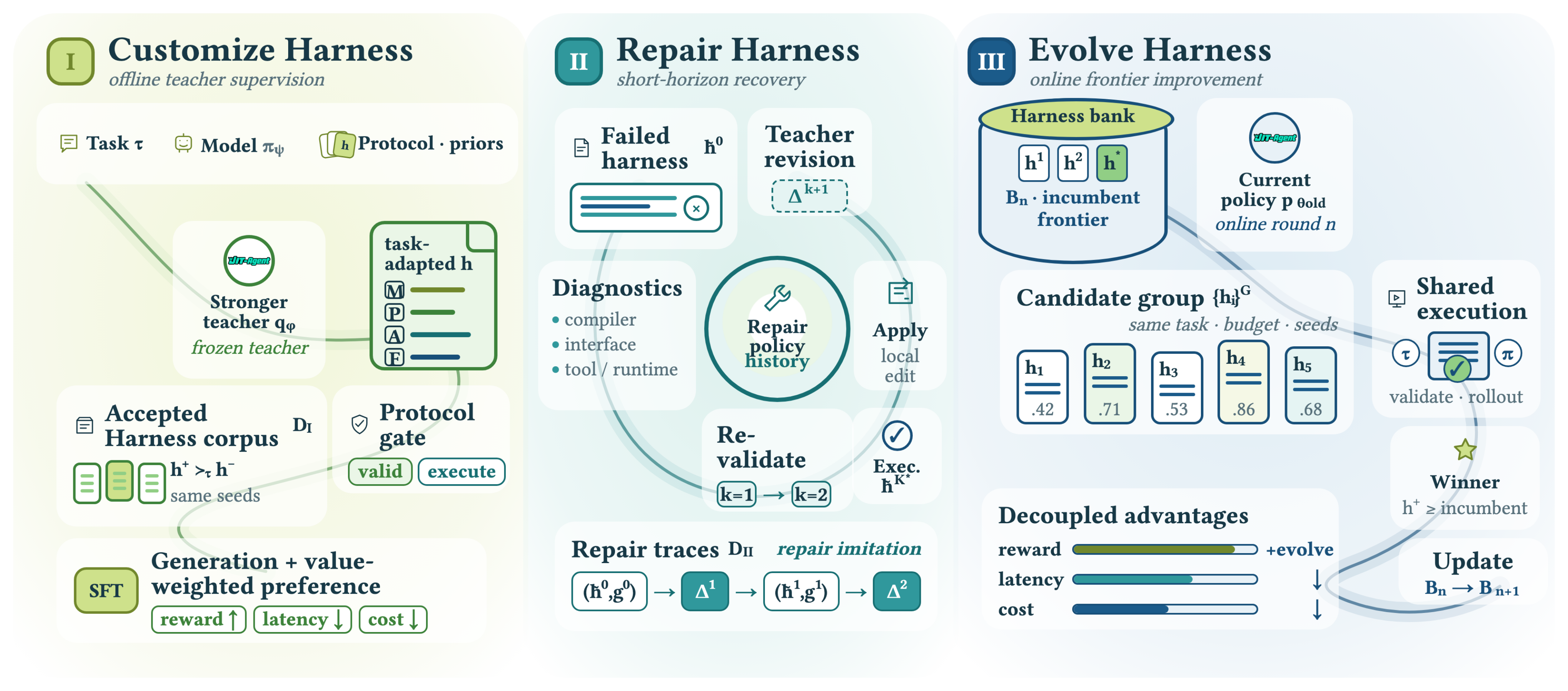}
   \vspace{-0.9em}
    \caption{\textbf{Training pipeline of \ourmethod.} Stage~I learns task-conditioned harness customization. Stage~II converts failed harnesses and execution diagnostics into bounded repair trajectories. Stage~III performs online evolution by comparing candidate harnesses against the incumbent bank, optimizing decoupled reward, latency, and cost advantages, and retaining frontier-improving designs.}
    \label{fig:training}
\end{figure}

\vspace{-0.5em}
\section{Method}\label{sec:method}
\vspace{-0.5em}
\subsection{Training Pipeline}\label{sec:training}
\vspace{-0.5em}

Training follows the inference-time lifecycle of \ourmethod: synthesize a task-conditioned harness, recover unstable generations, and then improve from stronger archive states online. Let $\boldsymbol{\tau}\sim\mathcal{D}_{\mathrm{task}}$ denote a training task, $\mathcal{C}_{\tau}$ its capability registry, and $\mathcal{E}_{\tau}$ a small reference context from the harness bank. The generation context is $\mathbf{c}_{\tau}=(\boldsymbol{\tau},\boldsymbol{\Pi},\mathcal{C}_{\tau},\mathcal{E}_{\tau})$, from which $p_{\theta}$ generates a harness $\mathbf{h}\in\mathcal{G}$. A frozen executor $\pi_{\psi}\sim\mathcal{M}$ evaluates task reward, latency, and monetary cost, while $\operatorname{Valid}_{\boldsymbol{\Pi}}(\mathbf{h};\boldsymbol{\tau},\pi_{\psi},\mathcal{C}_{\tau})\in\{0,1\}$ checks protocol and execution validity and returns structured diagnostics upon failure. The three stages optimize this generator through offline supervision, repair-trajectory learning, and online policy improvement, respectively.

\vspace{-0.5em}
\paragraph{Stage I: Data Preparation.}\label{sec:scaffold_synthesis}
Stage I uses a frozen, stronger teacher $q_{\phi}$ to synthesize task-adapted harnesses under the fixed four-module protocol. For each task, $\mathcal{E}_{\tau}$ contains three scaffolds sampled from the task-type-matched subset $\mathcal{B}_0^{(d(\boldsymbol{\tau}))}$, where $d(\boldsymbol{\tau})$ denotes the task type. The teacher receives the resulting context $\mathbf{c}_{\tau}$, and its generation is retained only after protocol validation and execution checks. Training tasks come from existing agentic benchmarks and environments~\citep{wu2025webwalkerbenchmarkingllmsweb,shi2025taskcraftautomatedgenerationagentic,bai2026clawgymscalableframeworkbuilding,song2026envscalerscalingtoolinteractiveenvironments}, augmented with synthesized tasks.

\vspace{-0.5em}
\paragraph{Stage I: Training Setup.}
We first apply standard supervised fine-tuning to accepted teacher generations, teaching \ourmethod to map a task and its reference context directly to a protocol-compliant harness. Since executable harnesses can still differ substantially in quality and efficiency, we then compare candidates under the same backbone and evaluation seeds. A preference is retained only when reward improves, neither efficiency axis degrades, and at least one efficiency gain is strict:
\begin{equation}
\label{eq:stage1_value_gap}
\begin{aligned}
\mathbf{h}^{+} \succ_{\tau} \mathbf{h}^{-}
&\iff
r^{+} > r^{-}
\;\land\;
\ell^{+}\leq\ell^{-}
\;\land\;
\kappa^{+}\leq\kappa^{-}
\;\land\;
\bigl(\ell^{+}<\ell^{-}\;\lor\;\kappa^{+}<\kappa^{-}\bigr), \\
\Delta_{\mathrm{val}}(\boldsymbol{\tau};\mathbf{h}^{+},\mathbf{h}^{-})
&= \alpha_r (r^{+} - r^{-})
+ \alpha_{\ell}[\ell^{-}-\ell^{+}]_{+}
+ \alpha_{\kappa}[\kappa^{-}-\kappa^{+}]_{+},
\end{aligned}
\end{equation}
where $(r^{\pm},\ell^{\pm},\kappa^{\pm})$ are repeated-rollout averages for $(\mathbf{h}^{+},\mathbf{h}^{-})$, $[x]_{+}=\max(x,0)$, and $\alpha_r,\alpha_{\ell},\alpha_{\kappa}\geq0$ weight the reward, latency, and cost gaps. The resulting pairs supervise a standard DPO objective, weighted by $\Delta_{\mathrm{val}}$ and anchored to the frozen Stage-I SFT checkpoint. Combined with generation imitation, this objective preserves protocol-valid structure while favoring harnesses that are both more effective and more efficient.

\vspace{-0.5em}
\paragraph{Stage II: Data Preparation.}
Stage I optimizes for protocol compliance, but some generated harnesses still fail static or runtime validation. Rather than discarding them, Stage II pairs each failed harness $\widetilde{\mathbf{h}}^{(0)}$ with a structured diagnostic $\mathbf{g}^{(0)}$ that records compiler errors, interface mismatches, tool-call failures, or runtime exceptions. At repair round $k$, the teacher proposes a structured revision $\Delta^{(k+1)}$ from patch space $\mathcal{P}$; $\operatorname{Apply}$ updates the harness deterministically, after which validation produces the next diagnostic. We retain only trajectories that become executable within two rounds:
\begin{equation}
\label{eq:stage2_repair}
\begin{aligned}
\widetilde{\mathbf{h}}^{(k+1)}
&= \operatorname{Apply}\!\left(\widetilde{\mathbf{h}}^{(k)},\Delta^{(k+1)}\right), \qquad k\in\{0,1\},\\
K^{\star}
&= \min\Bigl\{k\in\{1,2\}:\operatorname{Valid}_{\boldsymbol{\Pi}}(\widetilde{\mathbf{h}}^{(k)};\boldsymbol{\tau},\pi_{\psi},\mathcal{C}_{\tau})=1\Bigr\}.
\end{aligned}
\end{equation}
Supervision is constructed only when $K^{\star}$ exists, focusing the corpus on realistic, locally recoverable failures rather than those requiring wholesale redesign.

\vspace{-0.5em}
\paragraph{Stage II: Training Setup.}
Stage II trains \ourmethod by teacher-forced imitation of the successful revisions $\{\Delta^{\star(k+1)}\}_{k=0}^{K^{\star}-1}$. Each revision is conditioned on $\mathbf{c}_{\tau}$ and the complete preceding harness--diagnostic history $\{(\widetilde{\mathbf{h}}^{(j)},\mathbf{g}^{(j)})\}_{j=0}^{k}$. Because $K^{\star}\le 2$, the model learns the short-horizon regime needed at deployment: use execution feedback to make a small number of high-leverage revisions that restore protocol-valid execution.

\vspace{-0.5em}
\paragraph{Stage III: Data Preparation.}\label{sec:harness_evolution}
Stage III treats \emph{test-time harness evolution} as a trainable capability: \ourmethod learns to propose harnesses that surpass prior designs and then become stronger references for subsequent tasks. We term the resulting online objective \textbf{Evolutionary Group-Decoupled Policy Optimization (Evo-GDPO)}, inspired by \citep{liu2026gdpogrouprewarddecouplednormalization}.
At online round $n$, we sample a task $\boldsymbol{\tau}$, retrieve prior high-quality designs $\mathcal{E}_{\tau,n}$ from the current bank $\mathcal{B}_n$, and form $\mathbf{c}_{\tau,n}=(\boldsymbol{\tau},\boldsymbol{\Pi},\mathcal{C}_{\tau},\mathcal{E}_{\tau,n})$. The model samples $G>1$ candidates independently from the current rollout policy; candidates and retrieved designs are evaluated under the same frozen executor $\pi_{\psi}\sim\mathcal{M}$, budget, and seeds. The highest-reward retrieved harness, with ties broken by lower latency and then lower cost, supplies the incumbent statistics $(b_r,b_{\ell},b_{\kappa})$.

\vspace{-0.5em}
\paragraph{Stage III: Training Setup.}
Every candidate is validated before execution; one that remains invalid after bounded repair receives the minimum task reward, while its repair latency and cost remain part of the measured efficiency. For executable candidate $i$, let $r_i$, $\bar\ell_i$, and $\bar\kappa_i$ denote reward, mean latency, and mean monetary cost over repeated rollouts. The reward channel is primary, whereas efficiency gains activate only when incumbent reward is preserved:
\begin{equation}
\label{eq:stage3_reward_channel}
\begin{aligned}
R^{\mathrm{rew}}_i = r_i + \lambda_{\mathrm{evo}}\,[r_i - b_r]_{+},\;
R^{\mathrm{lat}}_i = \mathbb{I}[r_i \ge b_r]\,[b_{\ell} - \bar{\ell}_i]_{+},\;
R^{\mathrm{cost}}_i = \mathbb{I}[r_i \ge b_r]\,[b_{\kappa} - \bar{\kappa}_i]_{+},
\end{aligned}
\end{equation}
where $\lambda_{\mathrm{evo}}\geq0$ controls the improvement bonus and $\mathbb{I}[\cdot]$ is the indicator function. Evo-GDPO normalizes the three channels independently within each candidate group before aggregation, preventing their scales from overwhelming one another:
\begin{equation}
\label{eq:stage3_decoupled_adv}
\begin{aligned}
A^{m}_{i}
&= \frac{R^{m}_{i} - \operatorname{mean}(\{R^{m}_{j}\}_{j=1}^{G})}
{\operatorname{std}(\{R^{m}_{j}\}_{j=1}^{G}) + \varepsilon_{\mathrm{num}}},
\qquad m\in\{\mathrm{rew},\mathrm{lat},\mathrm{cost}\}, \\
A^{\Sigma}_{i}
&= w_{\mathrm{rew}} A^{\mathrm{rew}}_{i}
+ w_{\mathrm{lat}} A^{\mathrm{lat}}_{i}
+ w_{\mathrm{cost}} A^{\mathrm{cost}}_{i},
\quad w_{\mathrm{rew}}>w_{\mathrm{lat}}+w_{\mathrm{cost}}.
\end{aligned}
\end{equation}
The nonnegative weights sum to one, and the stated inequality makes task reward dominant; $\varepsilon_{\mathrm{num}}>0$ stabilizes normalization. We further normalize $A_i^{\Sigma}$ across the training batch and use it in a standard clipped PPO objective with a token-level KL penalty to the frozen Stage-II checkpoint. Thus, unlike standard group-relative optimization, Evo-GDPO rewards candidates for overtaking prior designs and, when reward is preserved, for doing so more efficiently. After optimization, we update the harness bank conservatively: a candidate is retained only if it matches or exceeds the current reward frontier and then strictly improves at least one frontier dimension---either reward itself, latency, or cost. During training, this feedback updates both $\theta$ and $\mathcal{B}_n$; at deployment, $\theta$ remains frozen.

\vspace{-0.5em}
\subsection{Inference Architecture}
\vspace{-0.5em}

\ourmethod supports two inference modes: \textit{static inference} and \textit{streaming inference}; they differ in whether experience is discarded after the task or retained to support subsequent tasks.

\vspace{-0.7em}
\paragraph{Static inference.}
In this mode, we introduce a lightweight form of test-time scaling: \ourmethod generates $N$ harnesses in parallel, selects one of them, and executes only the selected harness. This increases candidate diversity without increasing the number of environment rollouts. The selected harness follows the same validation and bounded-repair procedure described above.

\vspace{-0.7em}
\paragraph{Streaming inference.}
Streaming inference mode is designed to carry useful experience forward across a sequence of tasks. For the $n$-th task $\boldsymbol{\tau}_n$, \ourmethod retrieves from the current bank $\mathcal{B}_n$, generates and selects a harness $\mathbf{h}^{\dagger}_n$, and executes it once. The resulting environment feedback is then used only to determine whether this experience should update the bank:
\begin{equation}
\label{eq:streaming_inference}
\begin{aligned}
\boldsymbol{\xi}_{n}
\sim \operatorname{Rollout}(\boldsymbol{\tau}_n,\pi_{\psi},\mathbf{h}^{\dagger}_{n},\mathcal{C}_{\tau_n};\boldsymbol{\Pi}), 
\mathbf{m}_{n}
&= \operatorname{Eval}(\boldsymbol{\xi}_{n}), \\
\mathcal{B}_{n+1}
= \operatorname{Update}_{\mathrm{III}}
(\mathcal{B}_{n};\boldsymbol{\tau}_n,\mathbf{h}^{\dagger}_{n},\mathbf{m}_{n}), 
\mathcal{E}_{\tau_{n+1},n+1}
&= \operatorname{Retrieve}
(\boldsymbol{\tau}_{n+1};\mathcal{B}_{n+1}),
\end{aligned}
\end{equation}
where $\mathbf{m}_{n}=(r_n,\bar\ell_n,\bar\kappa_n)$ contains reward, mean latency, and mean monetary cost. Following the Stage-III retention rule, $\operatorname{Update}_{\mathrm{III}}$ leaves $\mathcal{B}_n$ unchanged when the completed harness provides no admissible improvement. Otherwise, the retained harness becomes a potential reference for later tasks. Streaming inference therefore transfers prior experience through the evolving harness bank, without injecting environment feedback into the current rollout or updating model parameters.

\vspace{-0.5em}
\section{Experiments and Analysis}
\vspace{-0.5em}
\subsection{Experiment Setup}
\vspace{-0.5em}
\paragraph{Evaluation Benchmarks.}
We evaluate \ourmethod on nine benchmarks spanning four task types. \textbf{Deep Research} includes BrowseComp-Plus~\citep{chen2025browsecompplusfairtransparentevaluation} (accuracy), DeepSearchQA~\citep{gupta2026deepsearchqabridgingcomprehensivenessgap} (answer F1), and xBench-DeepSearch (xBench-DS)~\citep{chen2025xbenchtrackingagentsproductivity} (accuracy); \textbf{Daily Work} includes AgentIF-Oneday~\citep{chen2026agentifonedaytasklevelinstructionfollowingbenchmark} (normalized weighted rubric score) and PinchBench~\citep{pinchbenchBestModels} (average score); \textbf{Planning} includes the DeepPlanning-Shopping (cart match rate) and DeepPlanning-Travel (composite constraint score) subsets~\citep{zhang2026deepplanningbenchmarkinglonghorizonagentic}; and \textbf{Workspace} includes OfficeBench~\citep{wang2024officebenchbenchmarkinglanguageagents} and OdysseyBench~\citep{wang2025odysseybenchevaluatingllmagents} (task success rate for both). Together, these benchmarks cover evidence-intensive search, long-horizon instruction following, constraint-aware planning, and multi-application workspace execution. We report all metrics on a $0$--$100$ scale; higher is better.

\vspace{-0.5em}
\paragraph{Backbone Models.}
\ourmethod is trained based on Qwen3.6-27B. We primarily instantiate \ourmethod with {GLM-5.2}~\citep{zai2026glm52} and {DeepSeek-V4-Flash-Preview}~\citep{deepseek2026v4}. These backbones provide complementary operating points: GLM-5.2 is a strong open long-horizon model, whereas DeepSeek-V4-Flash emphasizes inference efficiency. To test the generalization, additional models from Qwen3.6, Mimo-V2.5 are also used.

\vspace{-0.5em}
\paragraph{Baselines.}
We use two complementary baseline groups. First, the broad comparison includes vanilla \textbf{Qwen3.7-Plus}~\citep{qwen2026qwen37plus}, \textbf{GLM-5.2}, \textbf{DeepSeek-V4-Flash/Pro} (both preview version), \textbf{Kimi K2.7 Code}~\citep{moonshot2026kimik27code}, \textbf{GPT-5.6-Sol}~\citep{openai2026gpt56}, and \textbf{Gemini 3.1 Pro/3.5 Flash}~\citep{google2026gemini31pro,google2026gemini35flash}. Second, holding the backbone fixed, we compare \ourmethod with five reusable agent harnesses: \textsc{Claude Code}~\citep{githubGitHubAnthropicsclaudecode}, \textsc{Codex}~\citep{openaiCodexCoding}, \textsc{OpenCode}~\citep{githubGitHubOpencodeaiopencode}, \textsc{Hermes}~\citep{nous2026hermes}, and \textsc{NanoBot}~\citep{hkuds2026nanobot}. The first group measures end-to-end competitiveness, while the second isolates the contribution of harness design from that of model weights.

\vspace{-0.5em}
\subsection{Main Results}
\vspace{-0.5em}

\newcommand{\resultblank}{\textemdash}

\begin{table}[!t]
\centering
\setlength{\tabcolsep}{3.2pt}
\renewcommand{\arraystretch}{1.12}
\caption{Main results across nine agentic benchmarks. Rows prefixed by \ourmethod use the same backbone as their vanilla counterpart but replace its default scaffold with a JIT-generated harness. Scores are reported on a $0$--$100$ scale; higher is better. Top-1, Top-2, and Top-3 results in each column are highlighted in soft green, yellow-green, and yellow, respectively; Top-1 is also bolded.}
\label{tab:main_results_vanilla}
\begin{adjustbox}{max width=\textwidth}
\begin{tabular}{lccccccccc}
\toprule
\textbf{Model} & \multicolumn{3}{c}{\textbf{Deep Research}} & \multicolumn{2}{c}{\textbf{Daily Work}} & \multicolumn{2}{c}{\textbf{Planning}} & \multicolumn{2}{c}{\textbf{Workspace}} \\
\cmidrule(lr){2-4}\cmidrule(lr){5-6}\cmidrule(lr){7-8}\cmidrule(lr){9-10}
 & \textbf{BC+} & \textbf{DSQA} & \textbf{xBench} & \textbf{AgentIF} & \textbf{PinchBench} & \textbf{Shop} & \textbf{Travel} & \textbf{Office} & \textbf{Odyssey} \\
\midrule
Qwen3.7-Plus & 70.5 & 78.0& 75.0 & 59.9 & 80.5 & 75.2 & 52.1 & 56.6& 67.7\\
GLM-5.2 & 72.0 & \toptwo{89.2}& 76.0 & 63.0 & \topthree{87.0} & 78.2 & 62.8 & 63.0& \topthree{75.3}\\
DeepSeek-V4-Flash & 68.1 & 76.2 & 70.1 & 58.4 & 81.7 & 59.1 & 54.8 & 61.0 & 71.0 \\
DeepSeek-V4-Pro & 71.4 & 72.4& 79.0 & 56.5 & 61.1 & 71.1 & 55.2 & 62.0 & 72.0 \\
Kimi K2.7 Code & 73.6 & 87.8 & 79.0 & 57.3 & 76.1 & 72.8 & 56.9 & \toptwo{65.8}& 69.9 \\
GPT-5.6 &  \toptwo{76.9} & 76.0& 81.0 & \toptwo{68.0}& 84.2 & \toptwo{83.7} & \topone{84.9} & \topthree{65.3}& 68.7\\
Gemini 3.1 Pro & 67.5 & 75.8& \topthree{83.0} & 60.1& 81.0 & 77.4 & \topthree{70.7} & 60.6& 74.0\\
Gemini 3.5 Flash & \topthree{75.0} & \topthree{88.0}& \toptwo{85.0} & \topthree{64.0}& 74.2 & 76.2 & 50.3 & 63.3& \toptwo{78.0}\\
\midrule
\ourmethod + GLM-5.2 & \topone{78.0} & \topone{93.9}& \topone{88.0} & \topone{69.9} & \topone{93.3} & \topthree{83.4} & \toptwo{83.0} & \topone{68.4} & \topone{78.7} \\
\ourmethod + DeepSeek-V4-Flash & 74.0 & 85.1 & 82.0 & 63.8 & \toptwo{92.9} & \topone{83.9} & 61.3 & 63.4 & 73.0 \\
\bottomrule
\end{tabular}
\end{adjustbox}
\end{table}

\Cref{tab:main_results_vanilla} compares JIT-generated harnesses with vanilla backbones and frontier models across nine benchmarks. Replacing the default scaffold improves all 18 matched backbone--benchmark pairs: the average score rises from $74.1$ to $81.8$ on GLM-5.2 ($+7.7$) and from $66.7$ to $75.5$ on DeepSeek-V4-Flash ($+8.8$). The largest gains appear on long-horizon planning, with $+24.8$ on DeepPlanning-Shopping for DeepSeek-V4-Flash ($59.1\!\rightarrow\!83.9$) and $+20.2$ on DeepPlanning-Travel for GLM-5.2 ($62.8\!\rightarrow\!83.0$); DeepSeek-V4-Flash also gains $11.9$ on xBench-DS and $8.9$ on DeepSearchQA. Despite using open backbones, the JIT-equipped systems lead eight of nine columns: \ourmethod{} + GLM-5.2 ranks first on seven benchmarks, including $93.9$ on DeepSearchQA, $69.9$ on AgentIF, and $93.3$ on PinchBench, while \ourmethod{} + DeepSeek-V4-Flash leads DeepPlanning-Shopping at $83.9$ and exceeds DeepSeek-V4-Pro on every reported benchmark by $8.7$ points on average. DeepPlanning-Travel is the sole exception, where the JIT-equipped GLM-5.2 remains within $1.9$ points of GPT-5.6. These results show that task-adaptive harnesses consistently raise the capability of a fixed backbone rather than merely benefiting a small subset of tasks.

\vspace{-0.5em}
\subsection{Comparison with Advanced Agent Harnesses}
\vspace{-0.5em}
\label{sec:advanced_harness_comparison}

\begin{table}[!t]
\centering
\small
\setlength{\tabcolsep}{2.2pt}
\renewcommand{\arraystretch}{1.08}
\caption{Controlled comparison with advanced agent harnesses. For each fixed backbone, we report task performance (Perf.), average token consumption per case in thousands (\#Tokens (K)), and API cost per case in USD. Performance is better when higher; tokens and cost are better when lower. The best value within each backbone group is bolded.}
\label{tab:advanced_harness_comparison}
\begin{adjustbox}{max width=\textwidth}
\begin{tabular}{llccccccccc}
\toprule
\textbf{Backbone} & \textbf{Harness} & \multicolumn{3}{c}{\textbf{DeepSearchQA}} & \multicolumn{3}{c}{\textbf{xBench-DS}} & \multicolumn{3}{c}{\textbf{AgentIF}} \\
\cmidrule(lr){3-5}\cmidrule(lr){6-8}\cmidrule(lr){9-11}
 & & \textbf{Perf. $\uparrow$} & \textbf{\#Tokens (K)} & \textbf{Cost $\downarrow$} & \textbf{Perf. $\uparrow$} & \textbf{\#Tokens (K)} & \textbf{Cost $\downarrow$} & \textbf{Perf. $\uparrow$} & \textbf{\#Tokens (K)} & \textbf{Cost $\downarrow$} \\
\midrule
\multirow{6}{*}{DeepSeek-V4-Flash} & \textsc{Claude Code} & 79.6 & 625 & \$0.088 & 75.0 & 559 & \$0.079 & \textbf{66.9} & 808 & \$0.114 \\
 & \textsc{Codex} & 77.8 & 760 & \$0.107 & 70.0 & 680 & \$0.096 & 58.5 & 870 & \$0.123 \\
 & \textsc{OpenCode} & 75.9 & 1,832 & \$0.258 & 65.0 & 1,157 & \$0.159 & 48.1 & 950 & \$0.135 \\
 & \textsc{Hermes} & 69.9 & 1,157 & \$0.163 & 72.0 & 1,254 & \$0.177 & 60.3 & 1,000 & \$0.142 \\
 & \textsc{NanoBot} & 80.4 & 924 & \$0.131 & 78.0 & 527 & \$0.075 & 53.1 & 1,034 & \$0.147 \\
 & \ourmethod & \textbf{85.1} & \textbf{400} & \textbf{\$0.066} & \textbf{82.0} & \textbf{212} & \textbf{\$0.039} & 63.8 & \textbf{476} & \textbf{\$0.097} \\
\midrule
\multirow{6}{*}{Qwen3.6-Flash} & \textsc{Claude Code} & 72.8 & 710 & \$0.140 & 58.0 & 650 & \$0.128 & 55.4 & 900 & \$0.177 \\
 & \textsc{Codex} & 68.5 & 980 & \$0.193 & 52.0 & 874 & \$0.172 & 34.2 & 839 & \$0.170 \\
 & \textsc{OpenCode} & 64.1 & 1,968 & \$0.384 & 44.0 & 1,300 & \$0.256 & 38.7 & 961 & \$0.217 \\
 & \textsc{Hermes} & 60.8 & 1,319 & \$0.256 & 36.0 & 1,155 & \$0.227 & 49.7 & 1,184 & \$0.223 \\
 & \textsc{NanoBot} & \textbf{74.2} & 892 & \$0.197 & 63.0 & 597 & \$0.119 & 43.5 & 950 & \$0.187 \\
 & \ourmethod & 70.3 & \textbf{464} & \textbf{\$0.095} & \textbf{70.0} & \textbf{300} & \textbf{\$0.069} & \textbf{58.3} & \textbf{394} & \textbf{\$0.078} \\
\bottomrule
\end{tabular}
\end{adjustbox}
\end{table}

\Cref{tab:advanced_harness_comparison} controls the backbone and compares six harnesses on DeepSeek-V4-Flash and Qwen3.6-Flash across DeepSearchQA, xBench-DS, and AgentIF, reporting performance, tokens, and API cost per case. \ourmethod achieves the best performance in four of six settings: it exceeds the strongest fixed harness by $4.7$ points on DeepSearchQA and $4.0$ on xBench-DS with DeepSeek-V4-Flash, and by $7.0$ on xBench-DS and $2.9$ on AgentIF with Qwen3.6-Flash. In the two remaining settings, it trails the best score by only $3.1$ and $3.9$ points while using substantially fewer tokens, confirming that no fixed harness dominates across tasks. More importantly, \ourmethod has the lowest token consumption and cost in all six settings, reducing per-case cost by $14.9$--$54.1\%$ relative to the cheapest fixed alternative, or $36.0\%$ on average. On DeepSeek-V4-Flash xBench-DS, it cuts tokens from $527$K to $212$K and cost from $\$0.075$ to $\$0.039$ while raising performance from $78.0$ to $82.0$; on Qwen3.6-Flash AgentIF, it reaches $58.3$ with $394$K tokens at $\$0.078$, compared with at least $839$K tokens and $\$0.170$ for fixed harnesses. The gains therefore arise from task-conditioned scaffolds rather than longer trajectories.

\vspace{-0.5em}
\subsection{Cost--Performance Pareto Frontiers}
\vspace{-0.5em}
\label{sec:pareto_frontier}

\begin{wrapfigure}{r}{0.54\textwidth}
    \vspace{-0.8em}
    \centering
    \includegraphics[width=\linewidth]{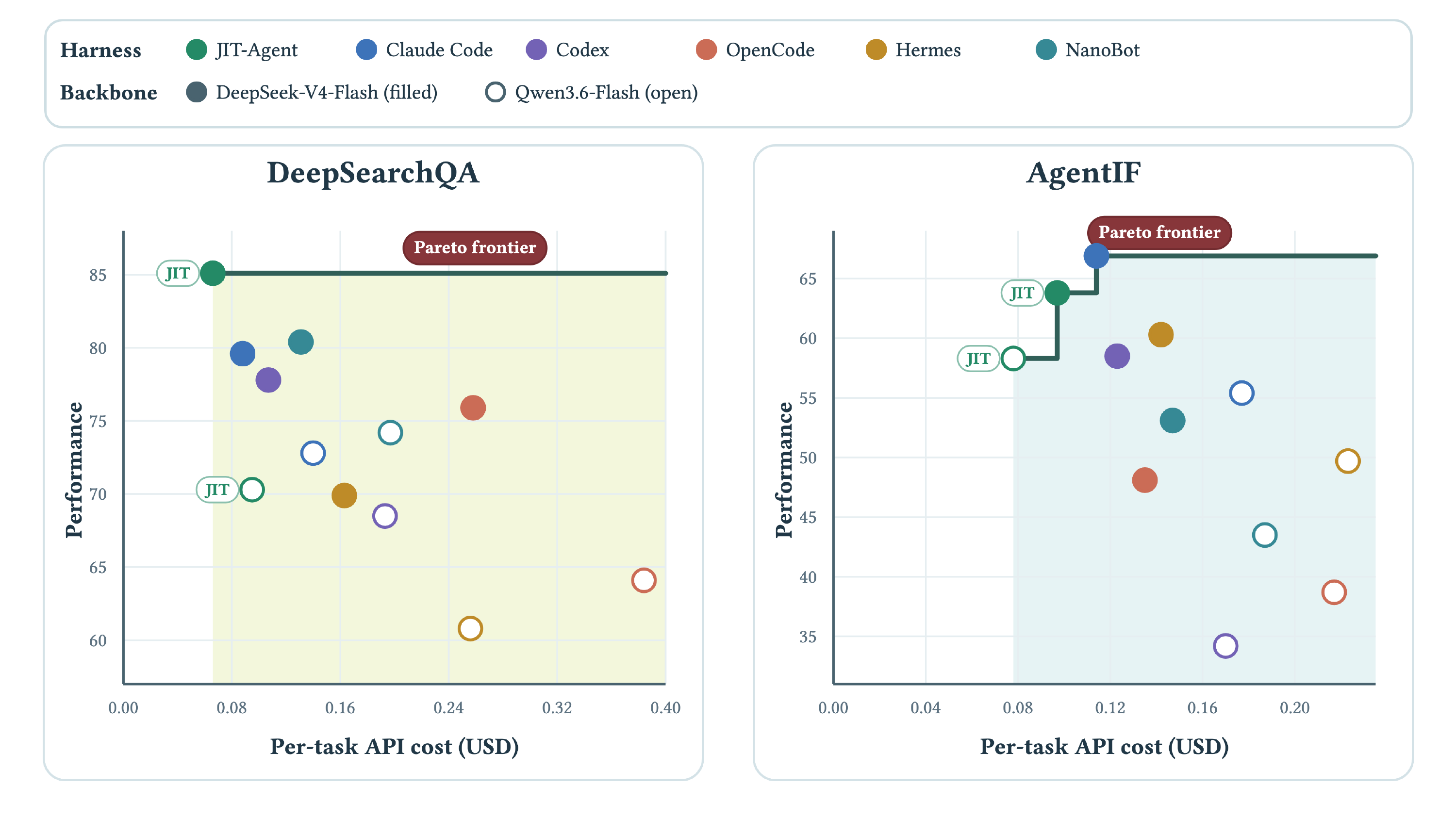}
    \vspace{-0.9em}
    \caption{\textbf{Cost--performance trade-offs on DeepSearchQA and AgentIF.}
    Marker color identifies the harness, while filled and open circles distinguish DeepSeek-V4-Flash and Qwen3.6-Flash. The horizontal axis is API cost per case in USD and the vertical axis is task performance. The dark-green step line traces the global Pareto frontier, and the pale yellow-green and blue-green regions contain pairings dominated by at least one Pareto-optimal point.}
    \label{fig:pareto_frontiers}
    \vspace{-0.8em}
\end{wrapfigure}

\Cref{fig:pareto_frontiers} visualizes the cost--performance geometry of the controlled comparison on DeepSearchQA and AgentIF. Each point is a backbone--harness pair from \Cref{tab:advanced_harness_comparison}; points closer to the upper-left corner are preferable.
On DeepSearchQA, DeepSeek-V4-Flash + \ourmethod is a strong Pareto-optimal operating point among the evaluated pairings, scoring $85.1$ at a per-case cost of $\$0.066$. Relative to the strongest fixed harness, NanoBot, it gains $4.7$ points while reducing cost by $49.6\%$ ($\$0.131\!\rightarrow\!\$0.066$). For Qwen3.6-Flash, the generated harness provides a lower-compute operating point: it scores $70.3$ at $\$0.095$, compared with NanoBot's $74.2$ at $\$0.197$, cutting cost by $51.8\%$ for a $3.9$-point trade-off.
AgentIF exhibits a different but complementary frontier. With DeepSeek-V4-Flash, \ourmethod reduces cost by $14.9\%$ relative to Claude Code ($\$0.097$ vs. $\$0.114$) while remaining within $3.1$ points of its performance ($63.8$ vs. $66.9$); both therefore lie on the frontier. With Qwen3.6-Flash, \ourmethod strictly dominates all fixed harnesses, improving the best score from $55.4$ to $58.3$ while lowering the minimum observed cost from $\$0.170$ to $\$0.078$. The frontier is therefore task-dependent, but the consistent leftward shift of the JIT points shows that task-adaptive harnesses improve inference efficiency rather than purchasing accuracy through longer trajectories.

\vspace{-0.5em}
\subsection{Generalization across Model Pairs}
\vspace{-0.5em}
\label{sec:model_pair_generalization}

\begin{figure}[!t]
    \centering
    \includegraphics[width=\textwidth]{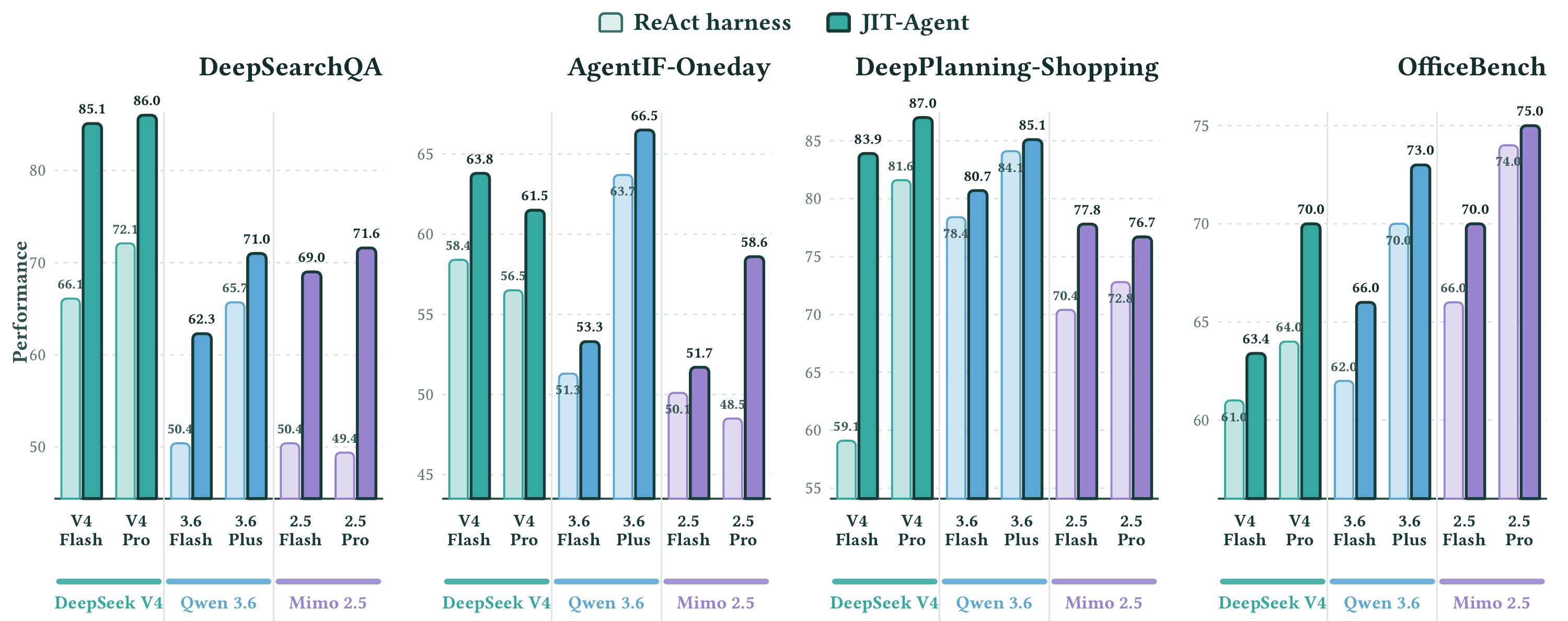}
    \caption{\textbf{JIT-generated harnesses consistently improve paired backbones over ReAct.}
    The four panels report DeepSearchQA, AgentIF-Oneday, DeepPlanning-Shopping, and OfficeBench. Within each panel, paired bars compare the same backbone under a fixed ReAct harness and a JIT-generated harness, with six backbones grouped by model family and scores shown above the bars. DeepSearchQA uses a 100-example subset; the other three benchmarks use 50-example subsets.}
    \label{fig:model_pair_jit_vs_react}
\end{figure}

To test transfer across model families and scales, we compare ReAct with JIT-generated harnesses on six fixed backbones from DeepSeek V4, Qwen3.6, and Mimo-V2.5 across four benchmarks. \Cref{fig:model_pair_jit_vs_react} shows that \ourmethod wins all 24 matched comparisons by $7.6$ points on average, with family-level gains of $10.2$, $4.0$, and $8.6$ points, respectively. DeepSearchQA benefits most ($+15.2$ on average), including $+22.2$ for Mimo-V2.5-Pro and $+19.0$ for DeepSeek-V4-Flash, while DeepPlanning-Shopping reaches a $+24.8$ gain. These consistent within-backbone improvements show that harness intelligence transfers across model families and variants rather than compensating for one backbone.

\vspace{-0.5em}
\subsection{Visualization of Generated Harnesses}
\vspace{-0.5em}
\label{sec:harness_visualization}

\begin{figure}[!tpb]
    \centering
    \includegraphics[width=\textwidth]{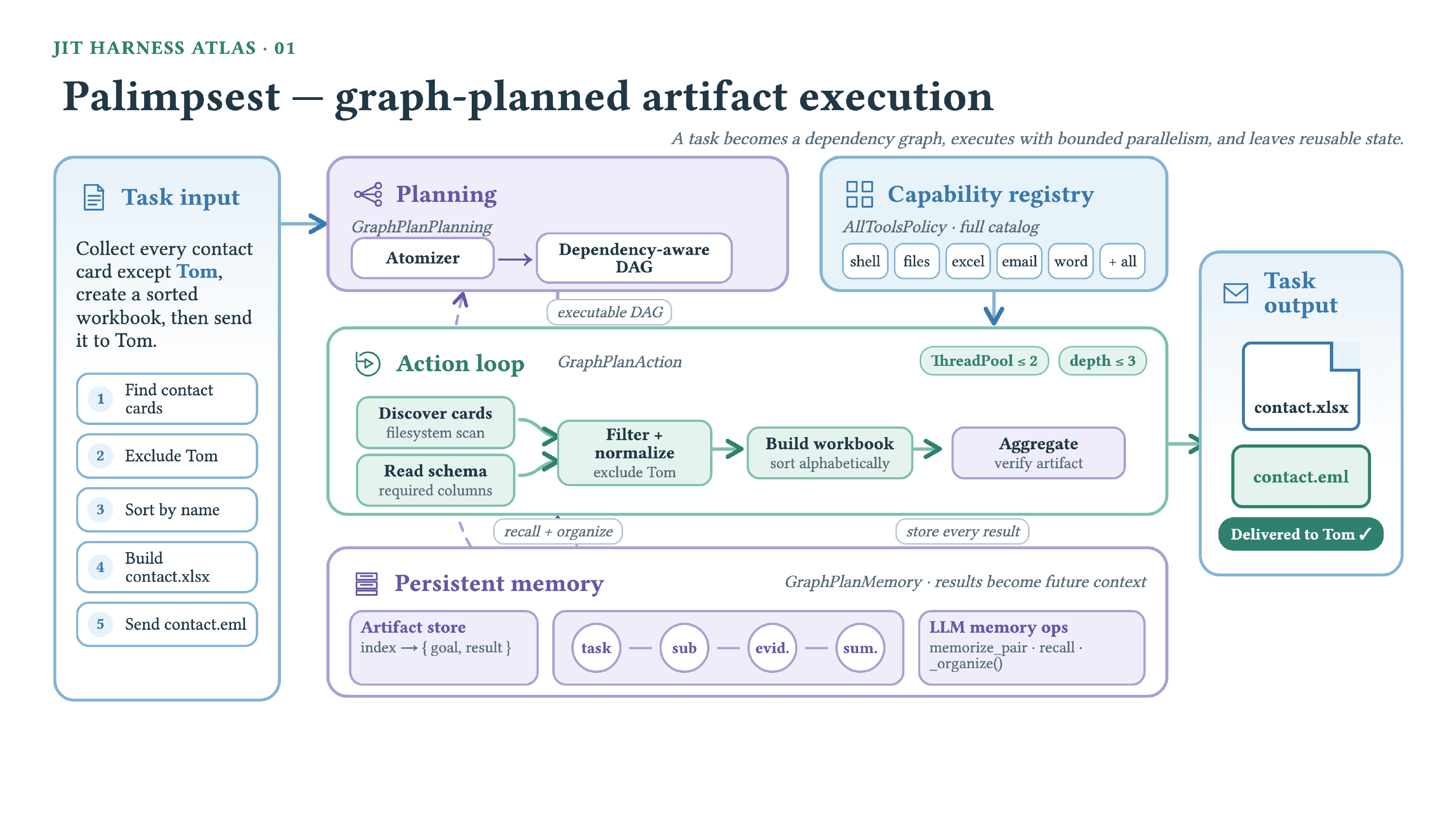}
    \vspace{-4em}
    \caption{\textbf{Palimpsest: graph-planned artifact execution.} \texttt{GraphPlanPlanning} converts the contact-processing request into a DAG; \texttt{GraphPlanAction} executes it with bounded width and depth, while \texttt{GraphPlanMemory} stores reusable artifacts and reasoning state.}
    \label{fig:harness-palimpsest}
    \end{figure}

To inspect the generated structures, \Cref{fig:harness-palimpsest} presents \textsc{Palimpsest}, a harness for discovering contact cards, filtering and normalizing records, constructing a workbook, and delivering it by email. It compiles these dependencies into a DAG: discovery precedes filtering, workbook construction waits for normalized records, and delivery follows artifact verification. \texttt{GraphPlanAction} executes ready nodes with bounded width and depth, while \texttt{GraphPlanMemory} preserves goals, evidence, and artifacts for downstream nodes. This task-shaped coordination of planning, memory, and execution goes beyond prompt variation and turns the request into a verifiable production pipeline; \Cref{sec:additional_harness_visualizations} shows complementary structures for recursive delegation, context folding, gated tools, phase-conditioned execution, evidence tracking, numerical state, and workspace repair.

\vspace{-0.5em}
\section{Conclusion}
\vspace{-0.5em}
We introduced \ourmethod, a dedicated model for synthesizing, repairing, and evolving task-conditioned agent harnesses at inference time. By expressing the harness through a composable four-module protocol and training the generator through customization learning, repair supervision, and Evo-GDPO, \ourmethod turns harness construction from a manually engineered, ahead-of-time artifact into a learned capability. Across deep research, daily work, planning, and workspace tasks, JIT-generated harnesses consistently strengthen their underlying backbones, compete with frontier models and advanced fixed harnesses, and improve cost--performance trade-offs. These results establish harness intelligence, \textit{i.e.}, the capacity to adapt the operational scaffold through which a model acts, as a trainable and transferable source of agent capability beyond model weights alone.

\vspace{-0.5em}
\section*{Contributors}
\vspace{-0.5em}
Core Contributors: Guibin Zhang, Leo Lu, Fangzhou Xie

Contributors: Kang Zhu, Junhao Wang, Zhifei Xie, Zhaochen Yu, Zihang Liu, Zhongxiang Sun, Qiankun Li, Yue Liao, Heng Chang, Xiaobin Hu, Qibing Ren

Corresponding: Wangchunshu Zhou, Chuanrui Hu, Yafeng Deng, and Shuicheng Yan

Affiliations: National University of Singapore,\;EverMind AI,\;Nanyang Technological University

\bibliographystyle{iclr2027_conference}
\bibliography{refs}

\clearpage
\appendix
\vspace{-0.5em}
\section{Harness Optimization Paradigms}
\vspace{-0.5em}
\label{sec:harness_optimization_appendix}
\Cref{tab:harness_optimization_comparison} organizes representative harness-optimization methods by when the harness is constructed and which parts of improvement are learned. Existing approaches primarily search for a durable harness ahead of deployment or edit that artifact online, whereas \ourmethod amortizes construction into a trained generator that synthesizes an instance-specific harness and learns both repair and evolution. The comparison is structural rather than a ranking of empirical performance: it clarifies which capabilities are native to the optimization procedure and which remain external search operations.

\begin{table}[!h]
\centering
\caption{\small \textbf{Harness optimization paradigms.} \emph{Construction} distinguishes harnesses found by ahead-of-time search, ahead-of-time harnesses subsequently edited with test-time feedback, and harnesses generated just in time for each task. \emph{Instance synthesis}, \emph{Harness model}, \emph{Learned repair}, and \emph{Online evolution} indicate whether a method directly synthesizes an instance-specific harness, trains the generator, learns repair from failed execution trajectories, and continues improving after deployment, respectively.}
\label{tab:harness_optimization_comparison}
\footnotesize
\setlength{\tabcolsep}{3pt}
\renewcommand{\arraystretch}{0.9}
\begin{tabular}{@{}>{\RaggedRight\arraybackslash}p{0.30\textwidth} >{\centering\arraybackslash}p{0.25\textwidth} *{4}{c}@{}}
\toprule
Method & Construction & \shortstack{Instance\\synthesis} & \shortstack{Harness\\model} & \shortstack{Learned\\repair} & \shortstack{Online\\evolution} \\
\midrule
AutoHarness~\citep{lou2026autoharness} & AOT (search) & \xmark & \xmark & \xmark & \xmark \\
Meta-Harness~\citep{lee2026metaharnessendtoendoptimizationmodel} & AOT (search) & \xmark & \xmark & \xmark & \xmark \\
AHE~\citep{lin2026agenticharnessengineeringobservabilitydriven} & AOT (search) & \xmark & \xmark & \xmark & \xmark \\
Adaptive AH~\citep{liu2026adaptiveautoharness} & AOT (test-time editing) & \xmark & \xmark & \xmark & \cmark \\
TTHE~\citep{nie2026tthe} & AOT (test-time editing) & \xmark & \xmark & \xmark & \cmark \\
RHI~\citep{lee2026recursiveharness} & AOT (test-time editing) & \xmark & \xmark & \xmark & \cmark \\
Harness-R1~\citep{shao2026harnessr1} & AOT (test-time editing) & \xmark & \cmark & \cmark & \cmark \\
\midrule
\textbf{\ourmethod (ours)} & \textbf{JIT} & \cmark & \cmark & \cmark & \cmark \\
\bottomrule
\end{tabular}
\end{table}

\vspace{-0.5em}
\section{Additional Results}
\vspace{-0.5em}
\label{sec:additional_results}
This section complements the aggregate comparisons in the main paper with the full streaming evaluation, which tests whether completed tasks can improve the harness bank for later tasks without changing the generator parameters.

\vspace{-0.5em}
\subsection{Test-Time Harness Evolution}
\vspace{-0.5em}
\label{sec:test_time_evolution}
\ourmethod is designed not only to generate a harness once, but also to improve its archive from execution feedback. We compare \emph{Static JIT}, where harness generations are independent, with \emph{Streaming JIT}, which retrieves and updates harnesses throughout the evaluation stream.

\begin{figure}[!t]
    \centering
    \includegraphics[width=\linewidth]{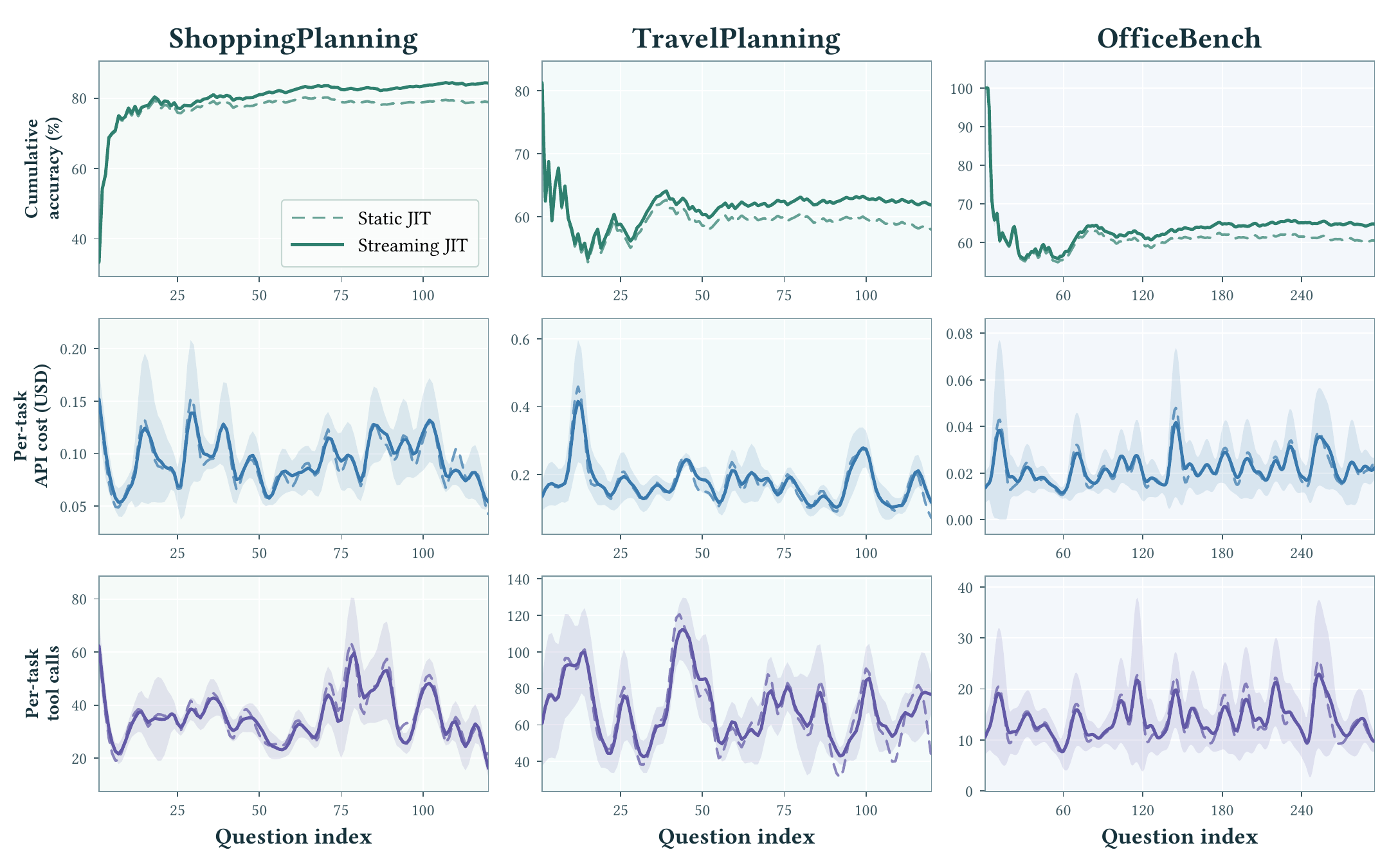}
    \caption{\textbf{Streaming test-time harness evolution across task streams.}
    Cumulative accuracy (top), per-task API cost (middle), and per-task tool calls (bottom) on DeepPlanning-Shopping, DeepPlanning-Travel, and OfficeBench. Dashed curves represent Static JIT, where task-specific harness generations are independent, while solid curves represent Streaming JIT, which continuously incorporates execution feedback as new tasks arrive. Shaded regions show local variation around the streaming trajectories. Streaming JIT finishes with higher cumulative accuracy on all three benchmarks, while API-cost and tool-use trajectories remain task-dependent and of broadly similar scale.}
    \label{fig:test_time_harness_evolution}
\end{figure}

\Cref{fig:test_time_harness_evolution} shows that Streaming JIT finishes above the static variant on DeepPlanning-Shopping, DeepPlanning-Travel, and OfficeBench. Across all three streams, the cumulative-accuracy advantage emerges as execution feedback accumulates and remains positive through the end of evaluation. The accompanying cost and tool-call traces suggest that the endpoint gains are not uniformly coupled to larger interaction budgets. This pattern is consistent with the objective of Evo-GDPO, which retains harnesses only when they advance the archive frontier.

\vspace{-0.5em}
\section{Additional Generated Harness Visualizations}
\vspace{-0.5em}
\label{sec:additional_harness_visualizations}
\setcounter{figure}{0}
\renewcommand{\thefigure}{\thesection.\arabic{figure}}
\renewcommand{\theHfigure}{\thesection.\arabic{figure}}
This appendix complements the representative case in \Cref{sec:harness_visualization} with nine additional harnesses generated under the same four-module protocol. The examples span bounded recursive research, constraint-heavy shopping and travel, phased web production, clue-driven research, evidence-sensitive retrieval, numerical analysis, and workspace manipulation. They are not variants of a single execution template: \ourmethod selects a state representation and control regime suited to the structure of each request, then instantiates planning, action, memory, and capability orchestration around that choice. Collectively, the cases make task-conditioned harness customization visible at the level of execution semantics rather than prompt wording.

\begin{figure}[!htbp]
    \centering
    \includegraphics[width=\textwidth]{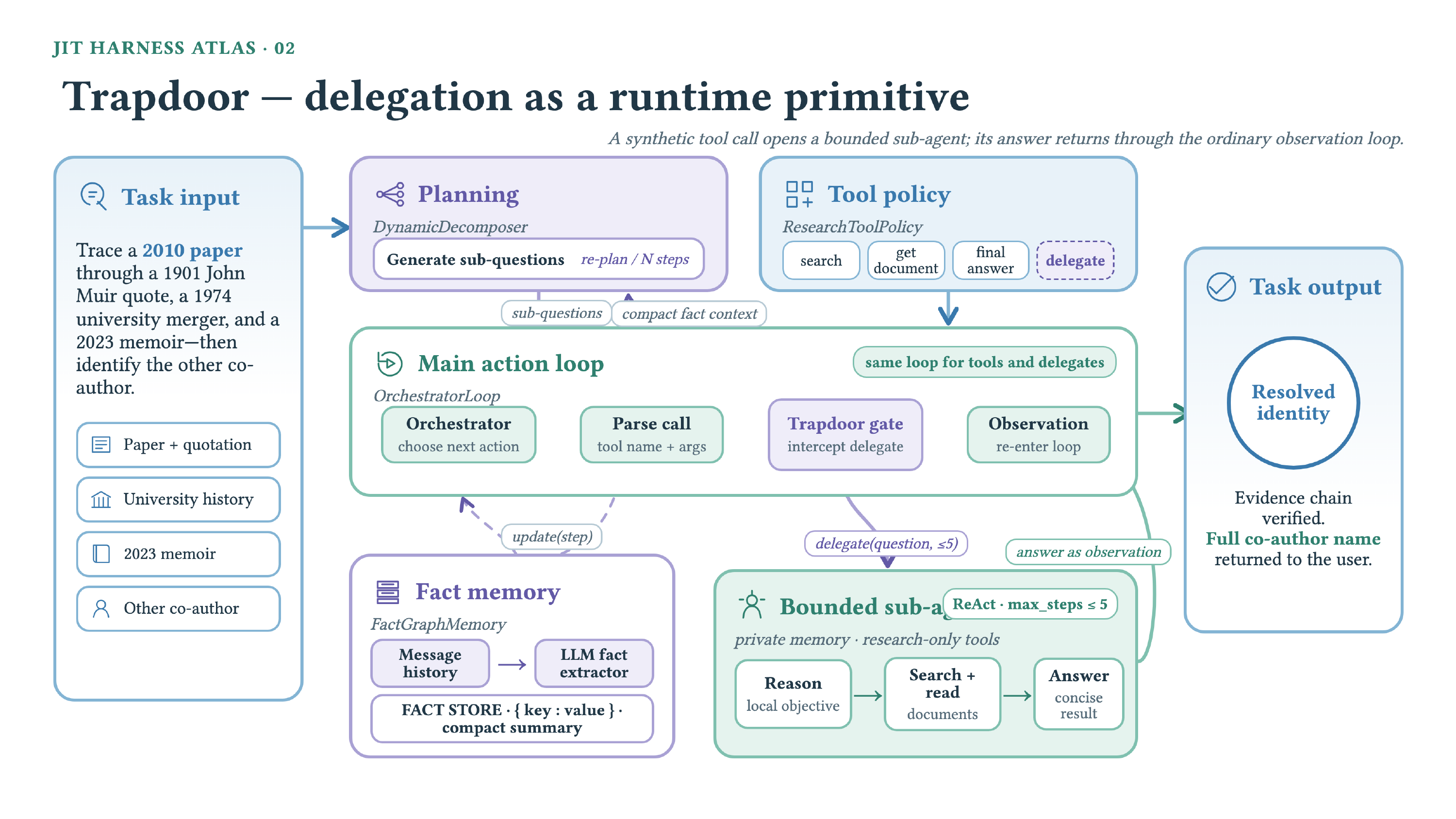}
    \vspace{-2em}
    \caption{\textbf{Trapdoor: bounded research behind a delegated tool call.} A synthesized \texttt{delegate} capability is intercepted by \texttt{OrchestratorLoop}, which runs a bounded subagent and writes extracted facts into \texttt{FactGraphMemory}.}
    \label{fig:harness-trapdoor}
\end{figure}

\vspace{-0.5em}
\paragraph{Deep research as bounded recursive delegation.}
\textsc{Trapdoor} addresses a multi-hop identity question whose clues span a historical quotation, a university merger, a memoir, and a research paper. Here a fixed DAG would commit too early to an uncertain evidence path, so the generated harness uses \texttt{DynamicDecomposer} to form and revise sub-questions and augments the research tool policy with a synthetic \texttt{delegate} capability. A \texttt{delegate} call opens a private research subagent with its own memory, research-only tools, and a five-step budget; the returned answer re-enters the parent loop as an ordinary observation. In parallel, \texttt{FactGraphMemory} extracts compact key--value facts from the growing history, allowing the parent to integrate evidence across branches without inheriting every subagent trace. Thus, the harness makes recursive research a runtime primitive specifically for a task whose uncertainty is resolved through branching evidence collection.

\begin{figure}[!htbp]
    \centering
    \includegraphics[width=\textwidth]{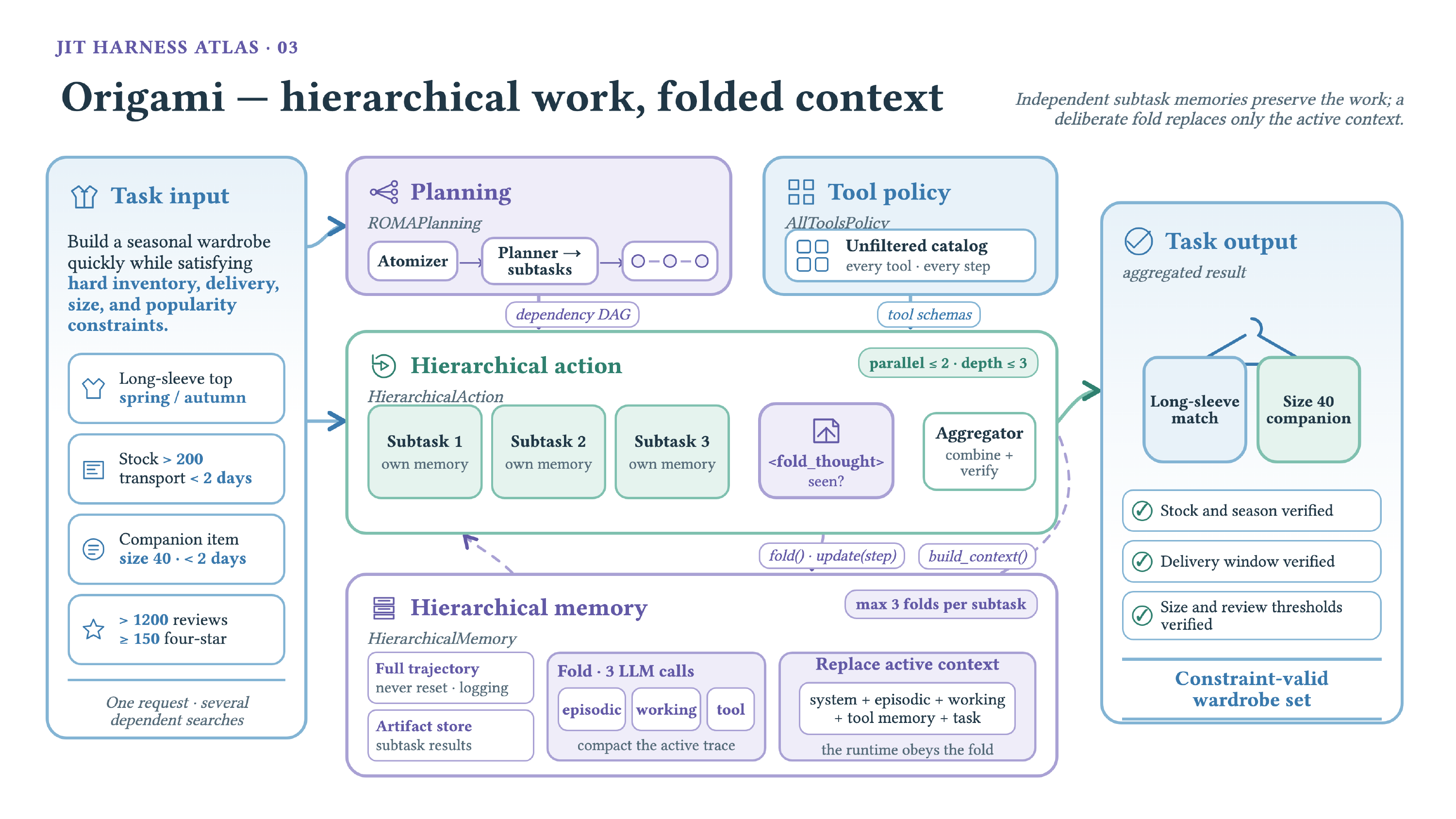}
    \vspace{-0.55em}
    \caption{\textbf{Origami: hierarchical work with folded context.} \texttt{ROMAPlanning} creates isolated subtasks; \texttt{HierarchicalMemory} retains their trajectories and artifacts while \texttt{fold\_thought} replaces only the active working context before aggregation.}
    \label{fig:harness-origami}
\end{figure}
\begin{figure}[!htbp]
    \centering
    \includegraphics[width=\textwidth]{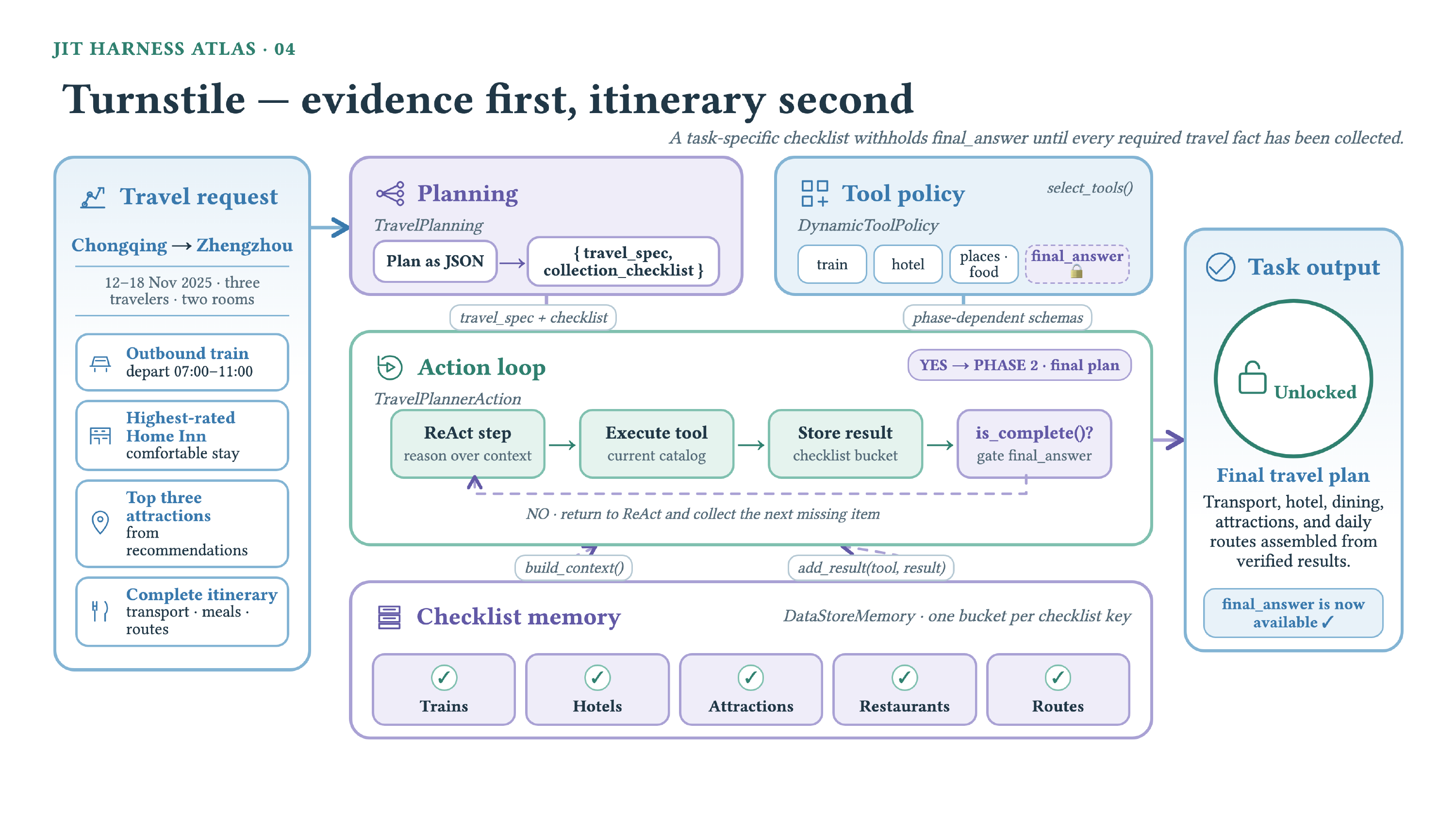}
    \vspace{-0.55em}
    \caption{\textbf{Turnstile: evidence first, itinerary second.} \texttt{TravelPlanning} emits a travel specification and checklist, \texttt{DataStoreMemory} tracks the required evidence buckets, and \texttt{DynamicToolPolicy} exposes \texttt{final\_answer} only after \texttt{is\_complete()} succeeds.}
    \label{fig:harness-turnstile}
\end{figure}

\Cref{fig:harness-origami,fig:harness-turnstile} illustrate two task-specific ways to control long-horizon execution through memory. \textsc{Origami} must assemble a seasonal wardrobe under coupled stock, delivery, size, review-count, and rating constraints. \texttt{ROMAPlanning} decomposes the request into dependent product searches, \texttt{HierarchicalAction} runs at most two branches in parallel to bounded depth, and each branch retains its own trajectory and artifacts. When a branch grows long, \texttt{fold\_thought} replaces only its active working context; completed subtask results remain available for the final constraint check. The harness therefore spends context on the unresolved portion of a combinatorial shopping task without discarding earlier product evidence.

\textsc{Turnstile}, by contrast, serves a travel request that specifies dates, passengers, rooms, a departure window, a highest-rated hotel, attractions, meals, and daily routes. The generated \texttt{TravelPlanning} module compiles these obligations into a typed \texttt{travel\_spec} and collection checklist, and \texttt{DataStoreMemory} assigns a dedicated bucket to each evidence class. \texttt{DynamicToolPolicy} exposes search tools for the missing bucket but withholds \texttt{final\_answer} until \texttt{is\_complete()} succeeds. Here memory is not used primarily to compress a long trace; it acts as an executable coverage contract that prevents itinerary synthesis before the required travel facts have been collected.

\begin{figure}[!htbp]
    \centering
    \includegraphics[width=\textwidth]{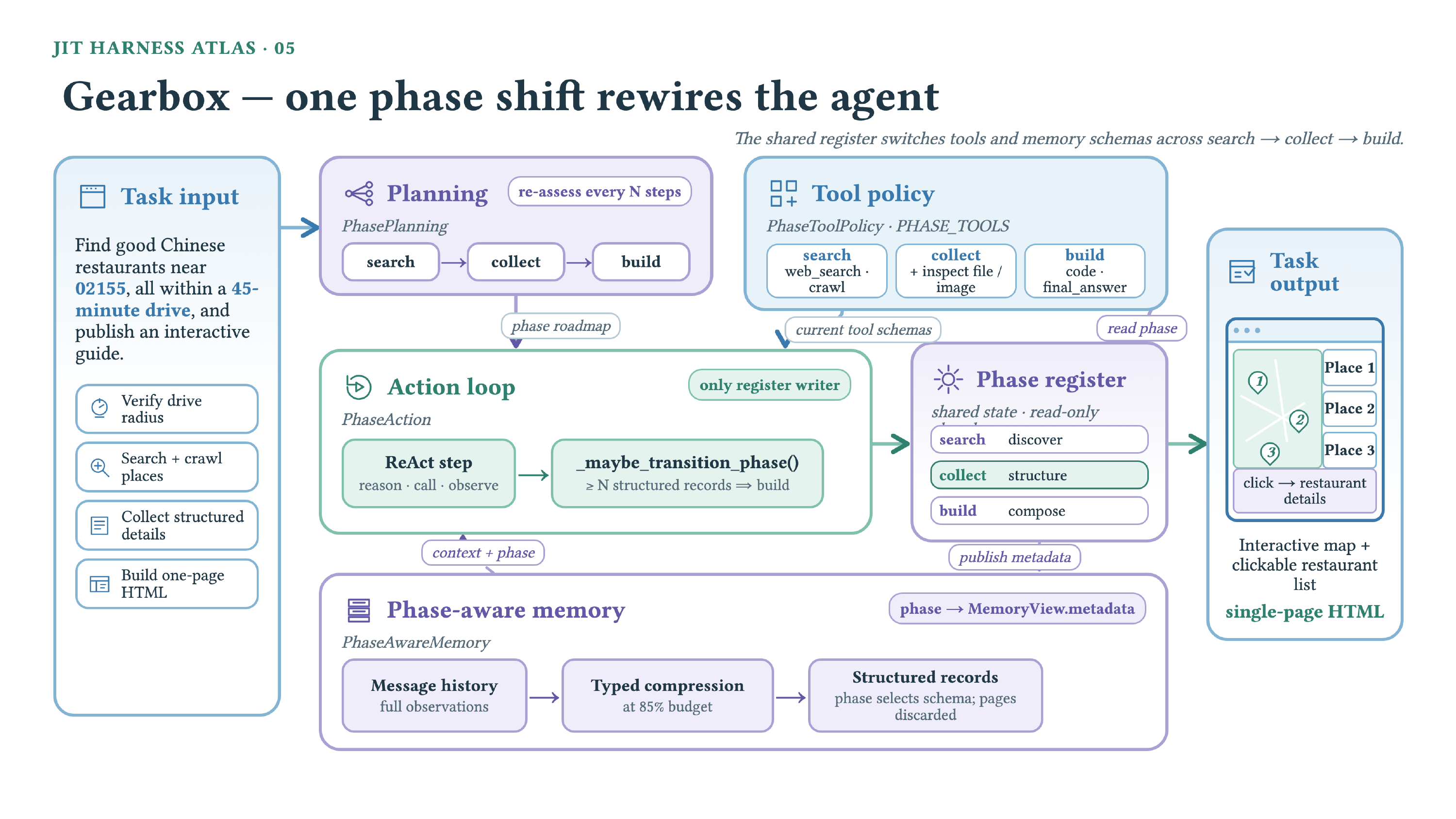}
    \vspace{-0.55em}
    \caption{\textbf{Gearbox: one phase shift rewrites the agent.} \texttt{PhaseAction} is the sole writer of a shared phase register; \texttt{PhaseToolPolicy} and \texttt{PhaseAwareMemory} read that state to switch both exposed capabilities and typed memory schemas.}
    \label{fig:harness-gearbox}
\end{figure}
\begin{figure}[!htbp]
    \centering
    \includegraphics[width=\textwidth]{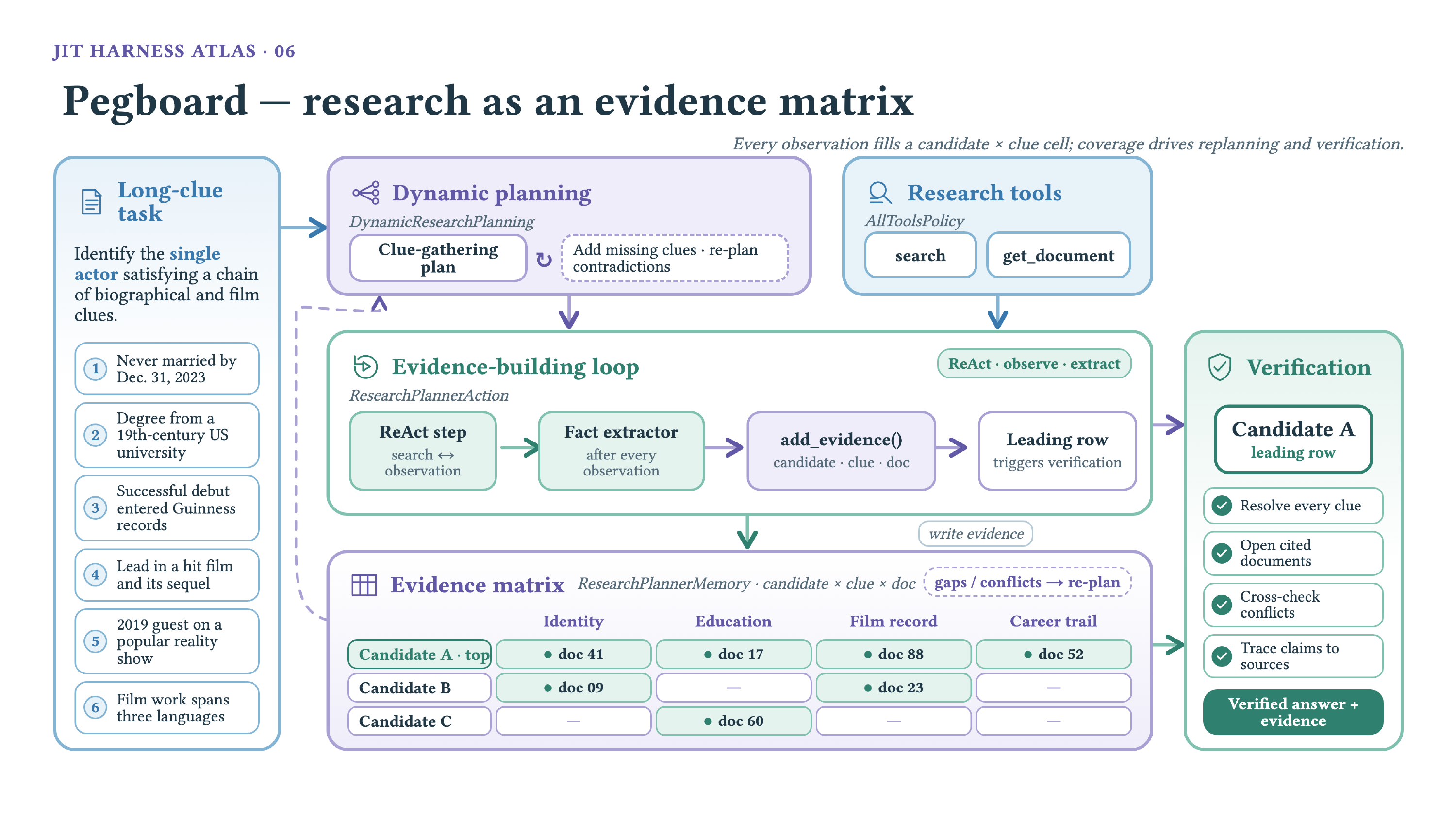}
    \vspace{-0.55em}
    \caption{\textbf{Pegboard: research as an evidence matrix.} Every observation is extracted into a candidate $\times$ clue cell with a document identifier; matrix coverage drives both \texttt{DynamicResearchPlanning} and the transition to source-grounded verification.}
    \label{fig:harness-pegboard}
\end{figure}

\Cref{fig:harness-gearbox,fig:harness-pegboard} show how explicit state can continuously reshape later decisions. \textsc{Gearbox} is generated to find Chinese restaurants within a specified driving radius and publish the results as an interactive one-page guide. Its \texttt{PhaseAction} is the sole writer of a shared \textit{search--collect--build} register; the transition to construction occurs only after enough structured restaurant records have been gathered. Both \texttt{PhaseToolPolicy} and \texttt{PhaseAwareMemory} read that register, switching from web discovery, to record inspection, to code and publication while compressing observations into the schema needed by the current phase. The model thus creates a phase machine because the task changes modality from evidence acquisition to artifact construction.

\textsc{Pegboard} instead tackles an identification problem defined by six biographical and film clues. It represents progress as a candidate $\times$ clue matrix in which every extracted claim retains its document identifier. Empty cells and contradictions drive \texttt{DynamicResearchPlanning} toward targeted searches, while a sufficiently supported row triggers a separate verification phase that reopens cited documents and resolves conflicts. Unlike a generic research transcript, this state makes coverage and source provenance directly actionable. The contrast with \textsc{Gearbox} shows that generated state may be a small phase variable or a task-shaped evidence structure, depending on what must govern the next action.

\begin{figure}[!htbp]
    \centering
    \includegraphics[width=\textwidth]{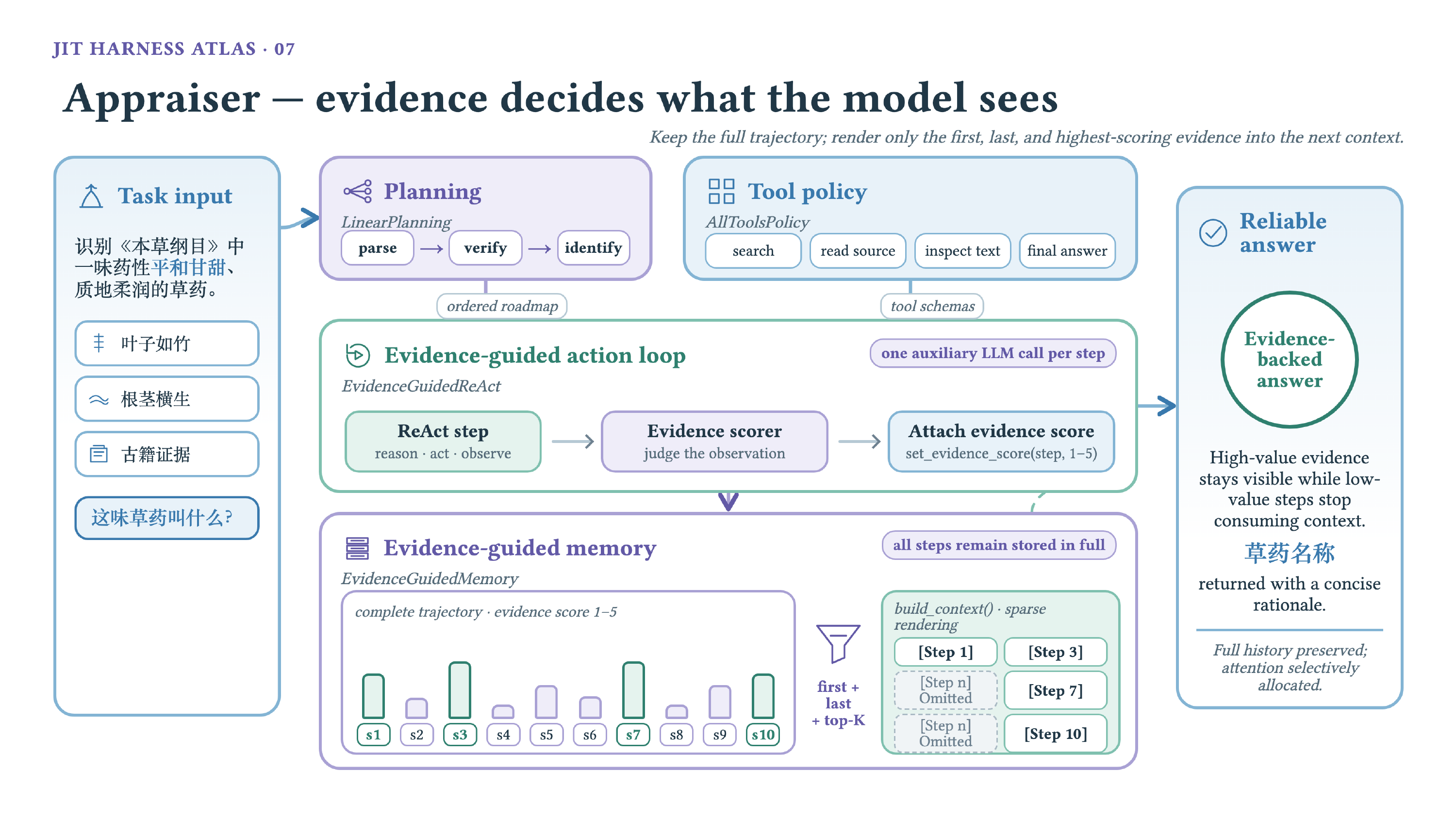}
    \vspace{-0.55em}
    \caption{\textbf{Appraiser: evidence decides what the model sees.} \texttt{EvidenceGuidedReAct} scores every step, while \texttt{EvidenceGuidedMemory} stores the full run but renders only the first, last, and top-$K$ observations into \texttt{build\_context()}.}
    \label{fig:harness-appraiser}
\end{figure}
\begin{figure}[!htbp]
    \centering
    \includegraphics[width=\textwidth]{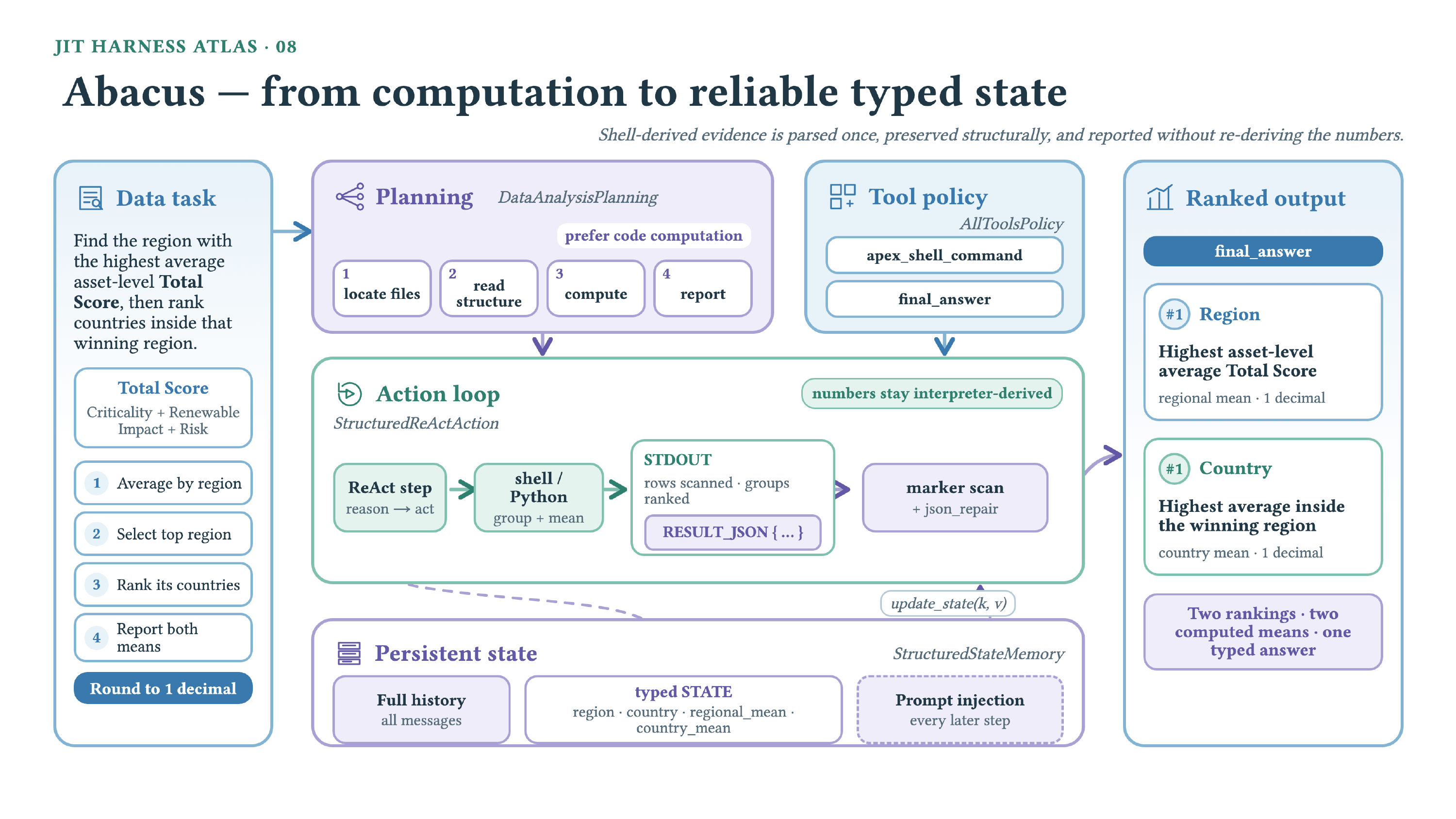}
    \vspace{-0.55em}
    \caption{\textbf{Abacus: from computation to reliable typed state.} \texttt{StructuredReActAction} executes shell or Python code, extracts a \texttt{RESULT\_JSON} payload from standard output, and updates \texttt{StructuredStateMemory} without asking the model to re-derive computed values.}
    \label{fig:harness-abacus}
\end{figure}

\Cref{fig:harness-appraiser,fig:harness-abacus} customize what subsequent reasoning receives from prior execution. \textsc{Appraiser} answers a classical materia-medica identification question by reconciling botanical traits with textual evidence. After every search or reading step, an auxiliary scorer assigns the observation an evidence value; \texttt{EvidenceGuidedMemory} keeps the complete trajectory for auditability but renders only the first, last, and top-$K$ observations into the next context. Low-value exploration therefore stops competing with the passages most useful for identification, while no evidence is erased from persistent history. This harness is tailored to a retrieval task in which the main bottleneck is evidence salience rather than tool choice.

\textsc{Abacus} is generated for a numerical analysis request that first compares regional averages and then ranks countries within the winning region. The harness deliberately moves aggregation out of language-model reasoning: \texttt{StructuredReActAction} invokes shell or Python, extracts a marked \texttt{RESULT\_JSON} payload from standard output, repairs its syntax if necessary, and writes the values into \texttt{StructuredStateMemory}. Region, country, regional mean, and country mean are then injected as typed state into every later step, so the final answer reports interpreter-derived quantities instead of re-deriving them from prose. Both cases preserve a full audit trail, but each renders the working view around its task's dominant reliability requirement---evidence selection for one and numerical fidelity for the other.

\begin{figure}[!htbp]
    \centering
    \includegraphics[width=\textwidth]{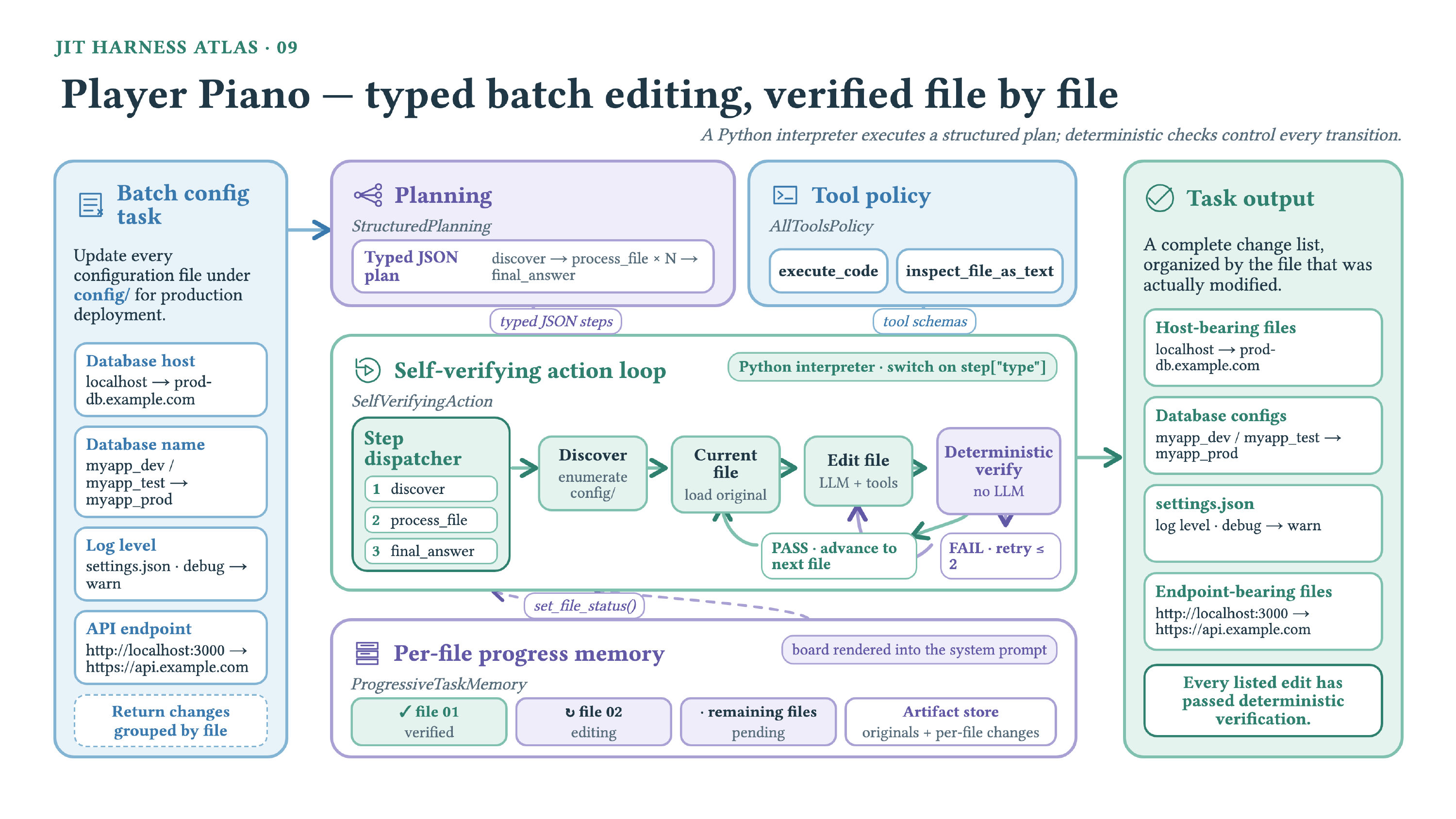}
    \vspace{-0.9em}
    \caption{\textbf{Player Piano: typed batch editing, verified file by file.} A Python dispatcher interprets \texttt{StructuredPlanning} steps, \texttt{SelfVerifyingAction} checks each edit without an LLM, and \texttt{ProgressiveTaskMemory} maintains a per-file board and artifact store.}
    \label{fig:harness-player-piano}
\end{figure}
\begin{figure}[!htbp]
    \centering
    \includegraphics[width=\textwidth]{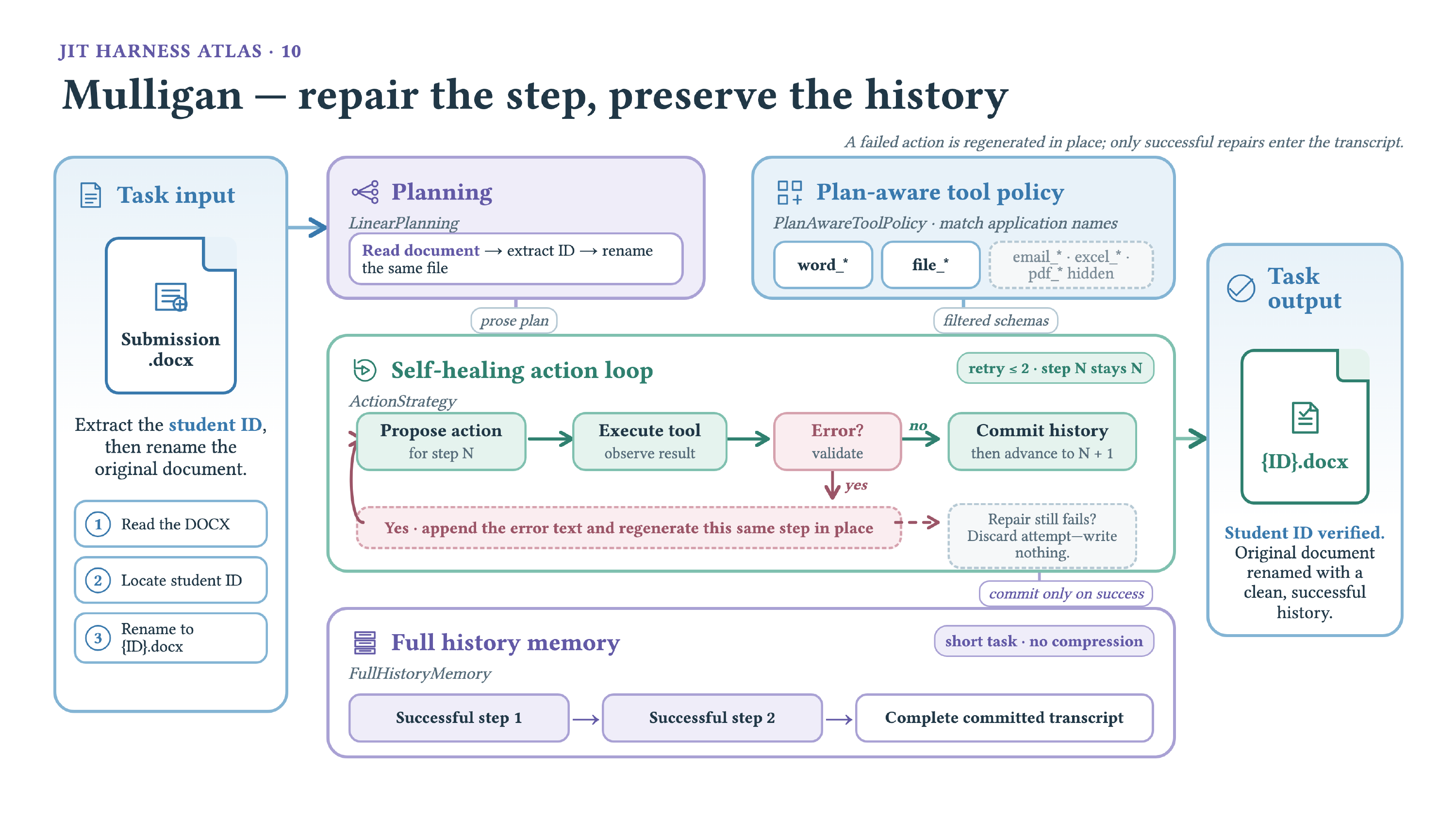}
    \vspace{-0.9em}
    \caption{\textbf{Mulligan: repair the step, preserve the history.} \texttt{PlanAwareToolPolicy} exposes only plan-relevant file and document tools; failed actions are regenerated at the same step, and \texttt{FullHistoryMemory} receives only the successful execution.}
    \label{fig:harness-mulligan}
\end{figure}

\Cref{fig:harness-player-piano,fig:harness-mulligan} specialize workspace execution at different scales by controlling when actions enter persistent history. \textsc{Player Piano} handles a batch migration of configuration files, including database, logging, and API-endpoint changes. \texttt{StructuredPlanning} emits typed \textit{discover}, \textit{process-file}, and \textit{final-answer} steps that a Python dispatcher executes; after each edit, a deterministic non-LLM check decides whether to advance or retry. \texttt{ProgressiveTaskMemory} renders a per-file board into the system prompt and retains both originals and accepted changes, preventing a large batch from losing track of which files have actually passed verification.

\textsc{Mulligan} is generated for a shorter document task: extract a student identifier from a DOCX file and rename that same file. Its linear plan needs no hierarchy or context compression, so the harness retains full successful history and exposes only Word and file operations while hiding unrelated email, spreadsheet, and PDF tools. If an action fails, the loop appends the error locally and regenerates the same step for at most two attempts; only a successful action is committed before the plan advances. The two harnesses therefore share a concern for reliable workspace mutation but realize it differently: one uses typed batch progress and per-file verification, whereas the other uses narrow tool exposure and transactional step repair. This task-dependent difference is precisely the behavior sought from \ourmethod: reliability mechanisms are generated to fit the structure and failure surface of the request rather than imposed as one universal scaffold.

\end{document}